\documentclass[hidelinks,onefignum,onetabnum]{siamart220329}
\pdfoutput=1

\usepackage{lipsum}
\usepackage{amsfonts}
\usepackage{graphicx}
\usepackage{epstopdf}
\usepackage{algorithmic}
\usepackage[english]{babel}

\newcommand{\firstrevnote}[1]{{\color{black}#1}}
\newcommand{\revnote}[1]{{\color{black}#1}}

\newsiamremark{hypothesis}{Hypothesis}
\crefname{hypothesis}{Hypothesis}{Hypotheses}
\newsiamthm{claim}{Claim}

\headers{Surrogate-based Autotuning for Randomized Sketching Algorithms}{Y. Cho, J. Demmel, M. Derezi\'{n}ski, H. Li, H. Luo, M. Mahoney, R. Murray}

\title{Surrogate-based Autotuning for Randomized Sketching Algorithms in
Regression Problems\thanks{Corresponding author: James W. Demmel, \url{mailto:demmel@berkeley.edu}}
\funding{
This research was supported by the Exascale Computing Project (17-SC-20-SC),
a collaborative effort of the U.S. Department of Energy Office of
Science and the National Nuclear Security Administration. We used
resources of the National Energy Research Scientific Computing Center
(NERSC), a U.S. Department of Energy Office of Science User Facility
operated under Contract No. DE-AC02-05CH11231.
Research was partially funded by SLICE Lab industrial sponsors and affiliates Amazon, AMD, Apple, Google, Intel, and Qualcomm.
This work was also partially funded by an NSF Collaborative Research Framework: Basic ALgebra LIbraries for Sustainable Technology with Interdisciplinary Collaboration (BALLISTIC), under NSF Grant Nos. 2004235 and 2004763.}
}

\author{Younghyun Cho\thanks{\firstrevnote{Santa Clara University (younghyun.cho@scu.edu). Most work of this paper was done when Younghyun Cho was at University of California, Berkeley.}}
\and James W. Demmel\thanks{University of California, Berkeley (demmel@berkeley.edu, haoyunl2@berkeley.edu, mmahoney@stat.berkeley.edu, rjmurray@berkeley.edu)}
\and Micha\l~Derezi\'{n}ski\thanks{The University of Michigan (derezin@umich.edu)}
\and Haoyun Li\footnotemark[3]
\and Hengrui Luo\thanks{Lawrence Berkeley National Laboratory (hrluo@lbl.gov)}
\and Michael W. Mahoney\footnotemark[3]\hspace{1mm}\footnotemark[5]\hspace{1mm}\thanks{International Computer Science Institute}
\and Riley J.  Murray\footnotemark[3]\hspace{1mm}\footnotemark[5]\hspace{1mm}\footnotemark[6]
}

\usepackage{amsopn}

\ifpdf
\hypersetup{
  pdftitle={An Example Article},
  pdfauthor={}
}
\fi

\usepackage{xcolor}
\usepackage{babel}
\usepackage{calc}
\usepackage{mathtools}
\usepackage{url}
\usepackage{bm}
\usepackage{amsmath}
\usepackage{amssymb}
\usepackage{graphicx}
\usepackage{rotating}
\usepackage{esint}
\usepackage{soul}
\usepackage{multirow}
\usepackage{array}
\usepackage{bm}

\newcommand{\R}{\mathbb{R}}

\usepackage{stmaryrd}

\newcommand{\trans}{{\mathsf{T}}}

\DeclareMathOperator{\rank}{rank}

\usepackage{babel}
\usepackage{adjustbox}
\definecolor{brightmaroon}{rgb}{0.76, 0.13, 0.28}

\begin{document}

\maketitle

\begin{abstract}
Algorithms from Randomized Numerical Linear Algebra (RandNLA) are known to be effective in handling high-dimensional computational problems, providing high-quality empirical performance as well as strong probabilistic guarantees. 
However, their practical application is complicated by the fact that the user needs to set various algorithm-specific tuning parameters which are different than those used in traditional NLA.
This paper demonstrates how a surrogate-based autotuning approach can be used to address fundamental problems of parameter selection in RandNLA algorithms.
In particular, we provide a detailed investigation of surrogate-based autotuning for sketch-and-precondition (SAP) based randomized least squares methods, which have been one of the great success stories in modern RandNLA.
Empirical results show that our surrogate-based autotuning approach can achieve near-optimal performance with much less tuning cost than a random search (up to about \firstrevnote{7.6x} fewer trials of different parameter configurations).
Moreover, while our experiments focus on least squares, our results demonstrate a general-purpose \emph{autotuning pipeline} applicable to any kind of RandNLA~algorithm.
\end{abstract}



\section{Introduction}
\label{sec:Introduction}

Randomized Numerical Linear Algebra (RandNLA) concerns the use of randomness
as a computational resource to aid in solving large-scale linear algebra problems.
While one may view its origins as going back to the 1980's, with work by \cite{Dixon:1983:powerMethod}, the theoretical principles of modern RandNLA were not established until the mid 2000's with the works by  \cite{DMM06,Sarlos06,DMMS07_FastL2_NM10_}.
Just a few years later, in the wake of these important theoretical results, the first evidence was provided that RandNLA algorithms could outperform classical NLA algorithms on particular classes of problems, most notably highly-overdetermined least squares \cite{AMT:2010:Blendenpik} and low-rank matrix approximation \cite{HMT:2011}.
The effectiveness of these methods has since been demonstrated in scientific
computing \cite{CLB:2021:sparseCholImplement}, data analysis \cite{drineas2012fast},
and visualization \cite{wilkinson_distance-preserving_2020,liao2023efficient}, among other~areas. See \cite{Woodruff:2014,drineas2016randnla,RandNLA_PCMIchapter_chapter,MT:2020,DM21_NoticesAMS,RandLAPACK_book} for recent surveys and perspectives on the field.

RandNLA algorithms can achieve state-of-the-art performance, compared to their traditional deterministic counterparts.
However, their efficiency and reliability are influenced by various algorithmic parameters, and these parameters are different than those used in traditional NLA algorithms.
The problem of selecting (near-)optimal values for these parameters in practice is challenging and important.
Indeed, the development of standard workflows for challenges such as this is recognized as a priority research area in the final report of the DOE workshop on Randomized Algorithms in Scientific Computing \cite{RASC_tech_report}.

\subsection{The tip of the iceberg: \textit{sketching matrices} in randomized preconditioning algorithms for least squares}

In the context of RandNLA, a \textit{sketching matrix} is a linear dimension-reduction map sampled at random from some distribution.
When a sketching matrix is applied to a larger \textit{data matrix} it produces a smaller matrix called a \textit{sketch}.
There is a large literature on the design of distributions for sketching matrices so that sketches contain ``useful'' information from the data matrix, with high probability.
The meaning of ``useful'' here depends on the linear algebra problem at hand and the intended way that the sketch will be used.

This article focuses on the fundamental problem of overdetermined least squares.
That is, we consider an $m \times n$ matrix $\bm{A}$ with $m \gg n$ and a vector $\bm{b} \in \R^m$, and we want to solve
\begin{equation}\label{eq:LS}
\bm{\tilde{x}} = \operatornamewithlimits{argmin}_{\bm{x}\in\R^n}\|\bm{A}\bm{x} - \bm{b}\|_2^2.
\end{equation} 
We also focus on an algorithmic paradigm called \textit{sketch-and-precondition} (SAP), wherein a $d \times n$ sketch $\bm{\hat{A}} = \bm{S}\bm{A}$ with $d \gtrsim n$ is processed to obtain a preconditioner, which is used to \firstrevnote{improve the solution quality} \eqref{eq:LS} by \firstrevnote{the chosen} iterative method.

\begin{figure}
    \centering
    \includegraphics[width=\linewidth]{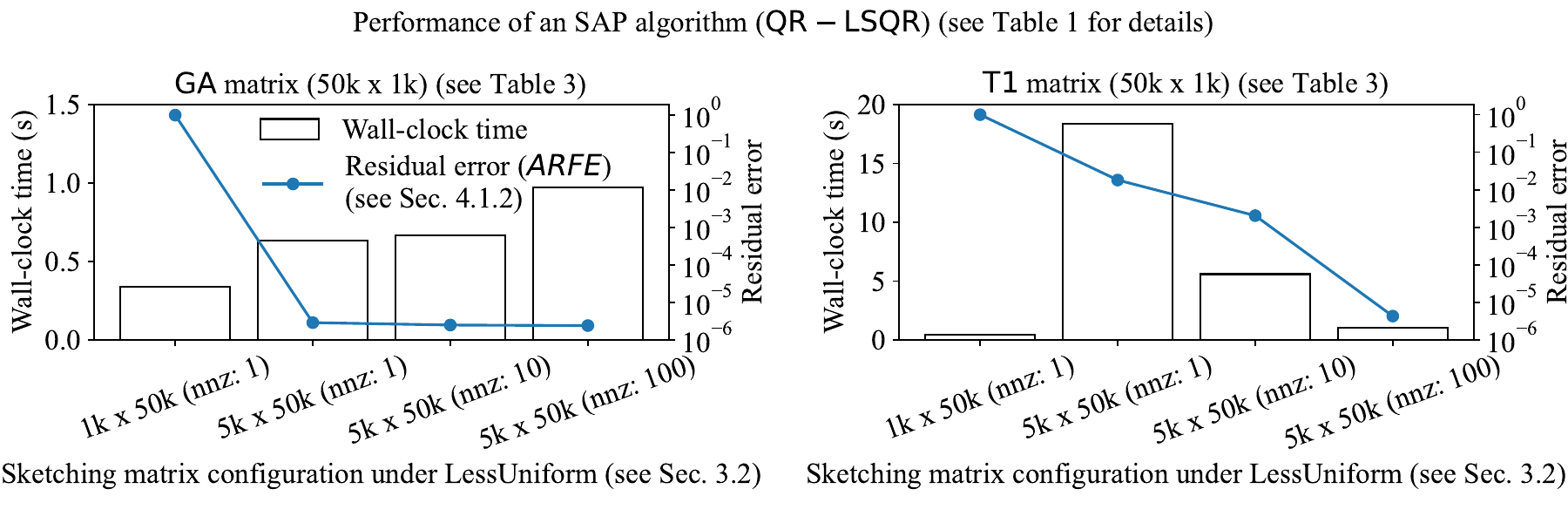}
    \vspace{-2em}
    \caption{
    \firstrevnote{Performance of a sketch-and-precondition (SAP) least squares algorithm with varying sketching matrices, i.e., with different sizes and/or different numbers of non-zeros in each row (nnz), for two different input matrices.}
    }
    \label{fig:analysis_motivation}
\end{figure}

As an illustration, Figure~\ref{fig:analysis_motivation} shows the performance of an SAP algorithm for various sparse sketching matrices. The distributions of these matrices are parameterized by the number of rows and the number of non-zeros in each row.
We can observe that the performance, in terms of both the wall-clock time of the solver and the resulting accuracy, can change significantly, depending on the sketching matrix.
The best performance of the solver comes only with choosing optimal parameters to construct the sketching~matrix, which already presents an interesting parameter tuning problem.
However, the question of how to \textit{simultaneously} tune parameters for the sketching matrix and how the sketch is processed is a far more challenging problem.

We should emphasize users need not go \firstrevnote{``blindly''} into the process of tuning a RandNLA algorithm.
For example, there are recommendations for parameter selection with probabilistic guarantees based on sophisticated mathematical analysis \cite{AMT:2010:Blendenpik,BG:2013,Cohen:2016:SJLTs,CW:2017:nnztime,DLPM21_newtonless_TR}.
The trouble is that these recommendations are often overly pessimistic \firstrevnote{and not adaptive to specific matrices} $\bm{A},\bm{b}$.
Therefore experts often rely on literature testimony for values of these parameters that are empirically considered ``safe,'' even if they are not theoretically justified.
However, even \textit{these} choices can lead to performance that is sub-optimal by an order of magnitude or more.
For example, in our extensive analysis in Section~\ref{sec:grid_search} with various randomized least squares solvers and sketching schemes, the optimal parameter configuration (found by an expensive grid search) \firstrevnote{achieves between 3.9x and 6.4x faster wall-clock times}, compared to a ``safe'' parameter configuration.
Of course, finding an optimal parameter configuration by exhaustive grid search is hardly a practical prescription for potential customers of RandNLA.

\subsection{Our contributions}
We propose an autotuning pipeline for RandNLA algorithms that is based on Bayesian optimization ideas.
In particular, as a case study, we provide a detailed analysis of performance tuning SAP-based randomized least squares methods, which have been among the great success stories in modern RandNLA.
We note up-front that a key strength of RandNLA lies in its agnosticism toward implementation details of subroutines used in broader algorithms. That is, RandNLA algorithms exhibit a high degree of \textit{modularity} \cite{Mah-mat-rev_BOOK,RandNLA_PCMIchapter_chapter,RandLAPACK_book}. 
Therefore while we focus on the problem of overdetermined least squares, the basic lessons from our work are applicable in other contexts, such as low-rank approximation. 
Indeed, our primary objective is to demonstrate a general \textit{workflow}, which can be applied towards autotuning RandNLA algorithms of very different flavors. 

\begin{figure}[t]
    \centering \includegraphics[width=\textwidth]{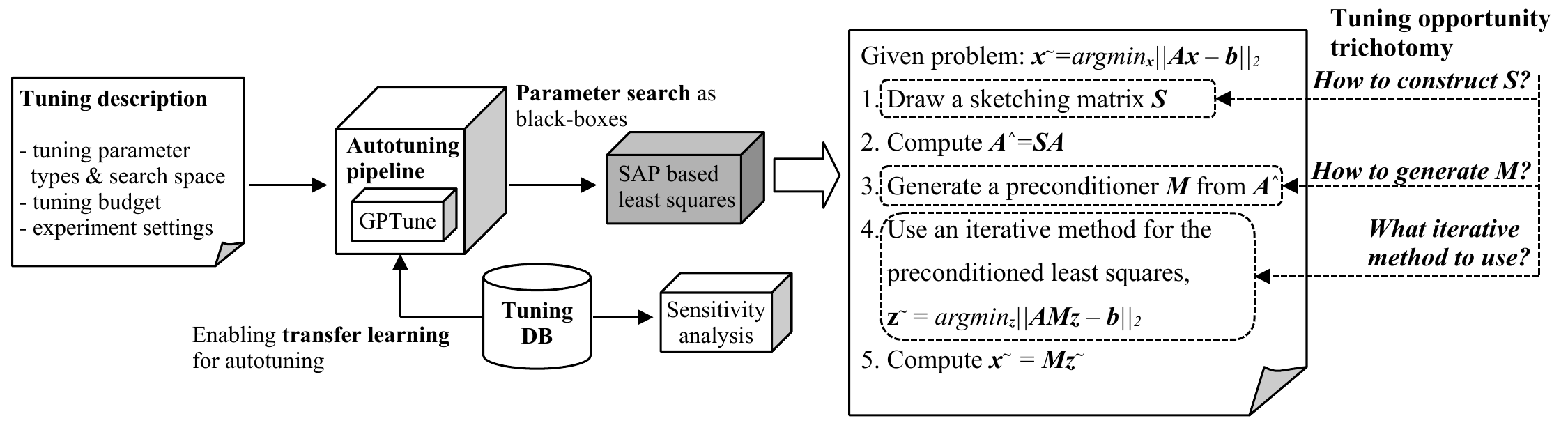}
    \vspace{-2em}
    \caption{Overview of our autotuning framework for SAP-based randomized least squares methods.
    The SAP methodology and its tuning opportunity trichotomy are detailed in Section~\ref{sec:sap_paradigm}.
    The actual parameter types, their search space, and the tuning algorithms are detailed in Section~\ref{sec:autotuning}.
    }
    \label{fig:tuning_overview}
\end{figure}

Figure~\ref{fig:tuning_overview} gives an overview of our autotuning framework.
The Bayesian tuning approach treats a given tuning problem as a black-box function, builds a surrogate performance model from measured performance results, and attempts to find optimal parameters using a limited number of trials.
The approach has garnered interest for tuning numerous problems that are difficult to model and costly to evaluate, such as hyperparameter tuning of machine learning models~\cite{Falkner:2018,JMLR:Hyperband} and performance tuning in high-performance computing (HPC) applications~\cite{balaprakash2018autotuning,Liu:2021:PPoPP,Cho:2023:IPDPS}.
As shown in the figure, we categorize the tuning opportunities of the SAP-based least squares methodology into three types (a \emph{tuning opportunity trichotomy}, see Section~\ref{sec:sap_paradigm}), and our autotuning pipeline tunes the SAP-based least squares solver as a black-box.

To the best of our knowledge, surrogate-based autotuning has not yet been studied well in the context of RandNLA algorithms.
Next we highlight some features of our approach and how these features factored into our experiments.

\paragraph{A stochastic surrogate model for a stochastic algorithm}
The stochastic nature of RandNLA results in variable performance, complicating the establishment of consistent performance measures for optimization.
Our autotuning framework uses an open-source surrogate modeling and tuning framework, GPTune~\cite{GPTuneUserGuide2022}.
GPTune uses Gaussian process (GP) regression to model the performance surface of the tuning problem and has demonstrated exceptional performance in diverse tuning settings \cite{Liu:2021:PPoPP,Cho:2023:IPDPS}.
GP methods are particularly well-suited to handle the uncertainties and variabilities introduced by the stochastic nature of RandNLA algorithms.

\paragraph{Verification of algorithm output}
Parameters in RandNLA algorithms affect multiple aspects of algorithm performance, including execution time and resulting accuracy.
As shown in Figure~\ref{fig:analysis_motivation}, for example, a certain sketching matrix configuration (when $\bm{S}$ is $1k \times 50k$ and has one non-zero per row) can prevent the SAP algorithm from accurately measuring the error of its approximate solution, which leads to it returning prematurely with an extremely inaccurate result.
Balancing performance metrics, specifically accuracy and speed, introduces further challenges \cite{liao2023efficient}.
For the tuning objective, we aim to minimize the execution time while attaining an acceptable range of resulting accuracy.
To this end, we compute obtained residual accuracy compared to  the solution found by a direct least squares solver and automatically filter out invalid parameter configurations based on their residual accuracy.

\paragraph{Transfer learning} The optimal parameter values for RandNLA algorithms can differ considerably based on the input problem (e.g., varying sizes/properties of input matrices).
Executing autotuning for each individual input problem would be prohibitively expensive.
To efficiently tune the randomized algorithm for various input matrices, we propose to leverage cross-input (matrix) information by employing the idea of ``transfer learning'' \cite{torrey2010transfer}.
In our autotuning context, transfer learning means that to tune on a given input matrix much more efficiently, we can leverage previously collected tuning information  from different matrices.
Several existing Bayesian autotuners~\cite{GPTuneUserGuide2022,Liu:2021:PPoPP,Menon_2020_IPDPS,Golovin:2017:KDD,Cho:2023:IPDPS} offer transfer learning algorithms (e.g., multitask learning using a Linear Coregionalization Model~\cite{Liu:2021:PPoPP}) that combines the GP models between the source and target datasets.
While we could employ these existing algorithms, we have observed that, for tuning SAP-based randomized least squares algorithms, the existing algorithms do not work well when handling categorical parameter search to select a specific algorithm and sketching operator.
To address this, and to further improve the tuning quality of transfer learning, our tuning pipeline employs a hybrid approach, where we explore the categorical space using a bandit learning algorithm \cite{auer2002using}, while applying the GP transfer learning techniques for other ordinal parameters.
We can further leverage GPTune's crowd-sourcing database~\cite{Cho:2023:IPDPS} which can facilitate such a transfer learning approach, by allowing multiple users (who might work on different input matrices) to share their data.
GPTune also provides data analysis techniques based on collected performance data,
such as sensitivity analysis, which estimates how sensitive the algorithm's performance is to each tuning parameter.
In this work, we perform such sensitivity analysis for each parameter of the RandNLA algorithms.

\paragraph{Numerical experiments}
We evaluate the effectiveness of our tuning pipeline using a set of synthetically generated input matrices with various properties,
\firstrevnote{as well as several real-world datasets.}
Our empirical results in Section~\ref{sec:experiments} demonstrate that our autotuning pipeline based on Bayesian optimization can attain near-optimal performance of the randomized least squares problem faster than simple tuning approaches such as grid and random search.
They also demonstrate that using transfer learning can further improve the tuning quality compared to non-transfer learning tuning, by attaining near-optimal performance using far fewer trials.
The results on transfer learning warrant special emphasis. 
\firstrevnote{For example, on a real-world matrix \textsf{Localization} (see Figure~\ref{fig:tuning_results_real_world}), with a given tuning budget of $50$ parameter configurations, the random search tuner achieves its best tuning result after evaluating $46$ parameter configurations; the result is on average $1.63$x slower than the peak performance we found from an expensive grid search.
In contrast, to attain the same (or better) performance, our surrogate-based approach using GPTune needs to evaluate only $13$ parameter configurations, and the transfer learning approach uses only $6$ parameter configurations.
These are $3.5$x and $7.6$x fewer number of evaluations than the random search, respectively.}

In addition to the autotuning experiments, Section~\ref{sec:grid_search} provides a detailed analysis of the performance of SAP algorithms for parameter configurations evaluated over a grid.
Of course, we do not recommend grid search as a method of parameter selection in RandNLA algorithms.
However, this detailed analysis reveals the ``true landscape'' of our tuning objective,
which helps explain the empirical results of the autotuner performance.

\revnote{
\subsection{Envisioned use cases}
\label{sec:envisioned_use_cases}
As discussed earlier, the optimal parameter configuration of the RandNLA algorithm depends on the given input matrix.
In this paper, we advocate using autotuning for RandNLA algorithms.
However, autotuning is an expensive process and is therefore often considered an offline approach in scenarios where tuned parameters can be used by many users/runs.
In this regard, one might wonder why we still want to use an autotuning approach for RandNLA algorithms, since different users may solve different input matrices, and autotuning needs to be done for each input matrix but is more expensive than the RandNLA solver itself.

In our envisioned use case, for example, if a user needs to run a RandNLA algorithm for a very large input problem (larger than the tested datasets in the present work), the user can first down sample the original input, apply our autotuning to efficiently explore the parameter space, and then run the RandNLA algorithm for the original, large input data, with a good parameter configuration.
Transfer learning may further accelerate the search by using previously collected samples from even smaller matrices, and we can allow sharing of performance samples among different users and use the previously collected performance samples.
Such specific case studies can be further researched in future work as well as several autotuning challenges we identified through this research (see Section~\ref{sec:conclusion}).

We emphasize that our work provides practical merits in performance optimization for the domain of RandNLA.
First, we discuss the empirical autotuning approach for the RandNLA domain for the first time to our best knowledge.
Second, we present an empirical tuning pipeline that uses GP regression along with an empirical evidence showing the potential of the surrogate-based autotuning approach for RandNLA.
}

\subsection{Outline}
The rest of this paper is organized as follows.
In Section~\ref{sec:surrogate_modeling}, we first provide the necessary background of the surrogate-based autotuning approach.
Then, we explain the SAP methodology for least squares and discuss its tuning opportunities in Section~\ref{sec:sap_paradigm}.
Section~\ref{sec:autotuning} explains the details of the autotuning techniques, and Section~\ref{sec:experiments} presents our experimental results to demonstrate the effectiveness of our autotuning framework.
We discuss related work in Section~\ref{sec:related_work} and conclude this paper in Section~\ref{sec:conclusion}.

\section{Surrogate modeling for performance tuning}
\label{sec:surrogate_modeling}

In this section, we provide background knowledge on the surrogate modeling techniques and their relevance to tuning RandNLA algorithms.

Analytical performance modeling is one of the widely used approaches for understanding and optimizing the performance of a certain algorithm.
Analytic performance models \cite{hong_complexity_1981,lo2014roofline} usually target certain performance metrics (e.g., flops or wall clock time) and provide mathematical models that are often human-readable under technical assumptions.
However, building an accurate analytic model for RandNLA algorithms is difficult, because of the complex interactions among parameter choices, and how they affect time per iteration and the number of iterations.

In contrast to the aforementioned analytic performance model, a ``black-box'' model, or a surrogate performance model only uses observed performance data to build surrogate models for certain metrics with respect to algorithm-related parameters \cite{GPTuneUserGuide2022,luo_nonsmooth_2021}.
Using a surrogate model with accurate predictions, the interaction between RandNLA's algorithm parameters and the performance can be well captured, analyzed and explored -- all without extensive background in RandNLA.
Surrogate models not only provide a data-driven approach for tuning algorithm parameters, but also they give insights into how these parameters affect the performance of the algorithm.

Among various surrogate modeling techniques, following \cite{gramacy2020surrogates} and \cite{shahriari_taking_2016}, we employ a Bayesian optimization scheme using Gaussian Process (GP) regression.
Bayesian optimization relies on sequential sampling that measures the application (at promising tuning parameter configurations) and builds a surrogate model using observed performance data.
The surrogate model is then used to choose the next tuning parameter samples.
This surrogate-based parameter selection brings us the following benefits.
\begin{itemize}
\item \textit{Efficiency.} We can achieve (near-)optimal parameter configurations using far fewer samples, compared to other simple tuning approaches such as exhaustive search or random search.
Compared to exhaustive search, the sequential sampling approach of Bayesian optimization evaluates the application for carefully chosen tuning parameter configurations guided by a surrogate model.
Unlike a random search, the surrogate-based approach can be proven to converge to optimal as the trial sample size increases \cite{shahriari_taking_2016}.

\item \textit{Transfer learning.}
We can transfer knowledge across different input problems, which can help tune on a given input problem if there are previously collected performance data for other input problems.
For example, a surrogate model trained for problems of one size can guide surrogate modeling for problems of another (larger) size.
Also, similar input problems may be correlated and have similar optimal parameters, providing a good warm start for the sequential sampling scheme in Bayesian optimization.

\item \textit{Sensitivity analysis.} We can deploy a sensitivity analysis on the surrogate performance model to help us quantify how changes to a given parameter are expected to affect the algorithm's performance, by computing sensitivity scores such as Sobol indices~\cite{sobol2001global}.
Compared to grid or random search, surrogate modeling can model performance using a smaller number of performance samples.
Therefore, the surrogate model-based sensitivity analysis is a computationally and time efficient approach.

\end{itemize}

\section{Sketch-and-precondition for least squares}
\label{sec:sap_paradigm}

To begin, recall that problem \eqref{eq:LS} features a tall $m \times n$ data matrix $\bm{A}$ and an $m$-vector $\bm{b}$.
The solution to this problem is $\bm{\tilde{x}} = \bm{A}^{\dagger}\bm{b}$, where $\bm{A}^{\dagger}$ is the Moore-Penrose pseudo-inverse of $\bm{A}$.

The SAP methodology for overdetermined least squares was introduced by \cite{RT:2008:SAP}. It was later extended with high-performance implementations and support for more general least squares problems through \textit{Blendenpik} \cite{AMT:2010:Blendenpik} and \textit{LSRN} \cite{MSM:2014:LSRN}. 
When applied to tall matrices, SAP methods offer accurate solutions at lower costs compared to direct methods.
This contrasts with \textit{sketch-and-solve} methods~\cite{DMM06,Sarlos06,DMMS07_FastL2_NM10_}, which are more suitable for computing low-accuracy approximations. 

\begin{algorithm}
\caption{Procedure of the SAP methodology}
\label{alg:sap_procedure}
\begin{algorithmic}[1]
\STATE Construct a $d \times m$ sketching matrix $\bm{S}$
\STATE Compute $\bm{\hat{A}} = \bm{S}\bm{A}$
\STATE Generate a preconditioner matrix $\bm{M}$ from $\bm{\hat{A}}$
\STATE Use an iterative method for the preconditioned least squares, \\ $\bm{\tilde{z}} = \operatornamewithlimits{argmin}_{\bm{z} \in \R^n} \|\bm{A}\bm{M}\bm{z} - \bm{b}\|_2^2$
\STATE Set $\bm{\tilde{x}} = \bm{M}\bm{\tilde{z}}$
\end{algorithmic}
\end{algorithm}

Algorithm~\ref{alg:sap_procedure} shows the modular subroutines of the SAP methodology.
SAP methods begin by constructing a wide $d \times m$ randomized sketching matrix $\bm{S}$ in the regime $d \gtrsim n$.
Then, SAP methods compute a $d \times n$ sketch $\bm{\hat{A}} = \bm{S}\bm{A}$.
The sketch is suitably factored to construct a preconditioner.
For example, we can compute the QR factorization (like Blendenpik~\cite{AMT:2010:Blendenpik}) or the SVD (like LSRN~\cite{MSM:2014:LSRN}) of $\bm{\hat{A}}$ to construct an $n \times n$ preconditioner matrix $\bm{M}$ such that $\bm{\hat{A}}\bm{M}$ is column-orthogonal.

The matrix $\bm{M}$ is used as a linear operator to right-precondition the original least squares problem.
If the distribution for the sketching operator $\bm{S}$ is chosen appropriately, then the orthogonality of $\bm{\hat{A}}\bm{M} = \bm{S}\bm{A}\bm{M}$ will (with high probability) result in $\bm{A}\bm{M}$ being \textit{nearly orthogonal}.
The near-orthogonality of $\bm{A}\bm{M}$ is valuable because it ensures rapid convergence of iterative methods for the preconditioned least squares problem $\bm{\tilde{z}} = \operatornamewithlimits{argmin}_{\bm{z} \in \R^n}\|\bm{A}\bm{M}\bm{z} - \bm{b}\|_2^2$.
The solution to that problem can be post-processed as $\bm{\tilde{x}} = \bm{M}\bm{\tilde{z}}$ to obtain an optimal solution for~\eqref{eq:LS}.

\subsection{Overview of tuning opportunities}
We categorize the central questions in the use of the SAP methodology into three categories, which we refer to as the {\emph{tuning opportunity trichotomy} \textsf{(TO1--3)}}.

\begin{enumerate}
    \item [\textsf{TO1:}] How should we construct the sketching matrix $\bm{S}$?
    \item [\textsf{TO2:}] How should we generate the preconditioner $\bm{M}$ from the sketched matrix $\bm{SA}$?
    \item [\textsf{TO3:}] What iterative method should we use for the preconditioned least squares, $\bm{\tilde{z}} = \operatornamewithlimits{argmin}_{\bm{z} \in \R^n} \|\bm{A}\bm{M}\bm{z} - \bm{b}\|_2^2$?
\end{enumerate}
To attain the best possible performance of the SAP paradigm, it is necessary to choose optimal settings for these questions.
For example, in Figure~\ref{fig:analysis_motivation} in Section~\ref{sec:Introduction}, we have shown that the best performance of an SAP method can only come with choosing optimal sketching configurations (\textsf{TO1}).
The preconditioner generation (\textsf{TO2}) and iterative method (\textsf{TO3}) are also the main factors influencing the performance of the SAP methodology, as we will show with a detailed analysis in Section~\ref{sec:experiments}.

\begin{table}[]
    \caption{SAP algorithm implementations considered in our autotuning study.}
    \label{tab:sap_algorithms}
    \centering
    \begin{tabular}{m{2.4cm}||m{2.8cm}|m{2.5cm}|m{3.5cm}}
        SAP algorithm & Preconditioner generation (\textsf{TO2}) & Iterative method (\textsf{TO3}) & Based on \\ \hline \hline
        \textsf{QR-LSQR} & QR & LSQR & Blendenpik~\cite{AMT:2010:Blendenpik}  \\ \hline
        \textsf{SVD-LSQR} & SVD & LSQR & LSRN~\cite{MSM:2014:LSRN}\\ \hline
        \textsf{SVD-PGD} & SVD & PGD & NewtonSketch~\cite{PW:2016:HessSketch,PW:2017:NewtSketch}\\
    \end{tabular}
\end{table}

Our autotuning study uses PARLA, which is a Python library for prototyping RandNLA algorithms \cite{PARLA}.
This library can run the SAP methodology with multiple choices regarding the tuning opportunity trichotomy.
For the sketching matrix generation (\textsf{TO1}), we consider a variety of \textit{sparse} sketching operators, which are particularly amenable to autotuning, thanks to the flexibility in the type and degree of sparsity that they employ.
We particularly consider two sparse sketching operators, SJLT and LessUniform, which will be detailed in Section~\ref{subsec:sketching_operators}.
For the preconditioner generation (\textsf{TO2}) and the iterative \firstrevnote{method} (\textsf{TO3}), this paper considers three SAP algorithms, based on three well-known ideas of Blendenpik~\cite{AMT:2010:Blendenpik}, LSRN~\cite{MSM:2014:LSRN}, and NewtonSketch~\cite{PW:2016:HessSketch,PW:2017:NewtSketch}.
Table~\ref{tab:sap_algorithms} shows the preconditioning and iterative methods for the three algorithms.
Note that these three algorithms use different combinations of preconditioner generation schemes (QR or SVD) and the iterative \firstrevnote{method} (LSQR or PGD, i.e., preconditioned gradient descent).
Note also that QR-based preconditioner generation with PGD as the iterative solver (what we might call \textsf{QR-PGD}) is not considered in this paper.
We provide the details of the preconditioner generation in Section~\ref{subsec:precond_gen} and the characteristics of the iterative methods in Section~\ref{subsec:determ_iter_solvers}.
Note that our algorithm implementations have subtle differences from the original works where the algorithms were first introduced; in Appendix~\ref{sec:sap_implementation}, we discuss how our SAP algorithm implementations are different compared to the original works.

\subsection{Details on sketching matrices}\label{subsec:sketching_operators}

Several types of distributions can be considered for the wide $d \times m$ sketching matrix~$\bm{S}$.
We consider two types of sparse sketching distributions, both of which have been well-studied in the~literature.

\begin{itemize}
    \item 
    \textbf{SJLT.}
    SJLT embeddings have independent columns, where each column is sampled by selecting $k$ indices from ${1,\ldots,d}$ uniformly without replacement. 
    These components are then set to $\pm 1/\sqrt{k}$ with equal probability, and the remaining values are set to zero. 
    The acronym ``SJLT'' stands for ``Sparse Johnson--Lindenstrauss Transform,'' the term first being used by \cite{DKS:2010:SJLT} in the theoretical computer science community.
    (One should note that this terminology has been used by different authors to refer to somewhat different embedding constructions.)
    Such sketching operators with $k$ non-zeros per column of $\bm{S}$ were first considered in the context of solving overdetermined least squares by \cite{CW:2013:CWTransform} (for a special case where $k=1$, known as the CountSketch). 
    The general variant, with $k\geq 1$, was later studied by \cite{NN:2013:OSNAPs} (under the name OSNAP) and \cite{Cohen:2016:SJLTs}. 
    For $k=d$, this sketching distribution recovers the classical dense subgaussian matrices with scaled random sign entries \cite{achlioptas2003database}.

    \item
    \textbf{LessUniform.}
    LessUniform embeddings have independent rows, where each row is sampled by selecting $k$ indices from ${1,\ldots,m}$ uniformly without replacement, and setting these components to $\pm \sqrt{m/kd}$ with equal probability.
    The name ``LessUniform'' derives from the concept of ``LEverage Score Sparsified embeddings'' (or LESS embeddings), originally proposed by \cite{DLDM20_sparse_TR}. 
    The original LESS embeddings are a family of sketching operators with $k$ non-zero entries per row of $\bm{S}$, where the placement of non-zeros within each row is determined by sampling from the leverage score distribution of $\bm{A}$.
    (Thus, the original LESS embeddings are a type of data-aware sketching operator; and, for $k=1$, they essentially reduce to leverage score sampling of the rows of $\bm{A}$ \cite{DMM06}.)
    For us, ``LessUniform'' refers to a simplified \textit{data-oblivious} variant of LESS embeddings~\cite{DLPM21_newtonless_TR,Der22_Algorithmic_TR}, which were proposed to avoid the cost of (approximately) pre-computing the leverage score distribution, while retaining the computational benefits of row-wise sparsity. 
    For $k=1$, this sketching distribution reduces to uniform sampling of the rows of $\bm{A}$, while for $k=m$, it reduces to a dense random sign matrix and has the same distribution as SJLT with $k=d$.
\end{itemize}

\noindent
Note that, in addition to sketch size $d$, both the SJLT and LessUniform sketching operators rely on a sparsity parameter $k$. 
However, crucially, the parameter plays a different role in each case, specifying the column-wise and row-wise sparsity of $\bm{S}$, respectively. 
Since the sketching operator is wide, i.e., $d\ll m$, a LessUniform matrix is much sparser than an SJLT matrix with the same parameters. 
This, in turn, is reflected both in the cost and the quality of the sketch, which makes it an interesting parameter space for autotuning.
Finally, since we focus on sparse sketching matrices, our parameterization does not include non-sparse distributions such as the subsampled randomized Hadamard transform (SRHT) \cite{AC:2006}. This is because varying the sparsity offers a richer parameter tuning landscape in comparison to SRHT, and moreover, our preliminary tests indicated that an SRHT-based approach would not improve upon sparse sketching operators.
Nevertheless, our tuning framework can also support tuning these and other sketching options, if the user wants to include more options.

\subsection{Details on preconditioner generation}
\label{subsec:precond_gen}

We obtain preconditioners using either the QR decomposition or the SVD of $\bm{\hat{A}}$. 
For the QR approach, the preconditioner is the $n \times n$ matrix $\bm{M} = \bm{R}^{-1}$ from $\bm{SA} = \bm{\hat{A}} = \bm{Q}\bm{R}$.
For the SVD approach, the preconditioner is the $n \times k$ matrix $\bm{M} = \bm{V}\bm{\Sigma}^{-1}$ from a compact SVD $\bm{\hat{A}} = \bm{U}\bm{\Sigma}\bm{V}^{\trans}$, where $k = \rank(\bm{\hat{A}})$.

SVD-based preconditioners have an advantage over QR-based preconditioners in that the former can be used to find \textit{minimum-norm} least squares solutions for rank-deficient problems.
Since we are interested in full-rank least squares problems, the primary question is one of \textit{efficiency}.
Here the SVD-based preconditioner will be more expensive to compute.
However, since $\bm{M} = \bm{V}\bm{\Sigma}^{-1}$ can be formed explicitly in $O(n^2)$ time given the SVD of $\bm{\hat{A}}$, it can be applied as a dense matrix-vector product.
Such an operation parallelizes better than the triangular solve required by the QR-based preconditioner.
\firstrevnote{In the QR-based approach, while there would be numerical issues with inverting $\bm{R}$, using it as a preconditioner would not have many numerical issues.}

Under the stated assumptions, there are no design considerations left that can affect the quality of the preconditioner.
This is because the simple assumption that $\bm{M}$ orthogonalizes $\bm{\hat{A}}$ completely determines the spectrum (i.e., singular values) of $\bm{A}\bm{M}$.
Specifically, the following fact holds.
\begin{proposition}\label{prop:precond_quality}
    Let $\bm{U}$ be an $m \times k$ matrix whose columns are an orthonormal basis for the range of $\bm{A}$.
    If $\rank(\bm{S}\bm{A}) = k$ and $\bm{M}$ is an $n \times k$ matrix for which $\bm{S}\bm{A}\bm{M}$ is orthogonal, then the spectrum of $\bm{A}\bm{M}$ is equal to that of $(\bm{S}\bm{U})^{\dagger}$.
\end{proposition}
Theorem 1 in \cite{RT:2008:SAP} (or rather, the proof of the same) implies as much under the assumption that $\bm{A}$ is full-rank.
A full proof of the claim in the general case is given in Appendix B of \cite{RandLAPACK_book}.

\subsection{Background on deterministic iterative solvers}
\label{subsec:determ_iter_solvers}

We consider preconditioned LSQR and preconditioned gradient descent (PGD)
for the deterministic iterative solvers in our study.\footnote{The latter solver can be viewed as the iterative method underlying NewtonSketch \cite{PW:2016:HessSketch,PW:2017:NewtSketch}, which is a method for solving structured convex optimization problems. See Appendix \ref{a:newtonsketch} for further discussion.}
When we apply these algorithms to right-preconditioned problems of the form
\begin{align}
    \bm{\tilde{z}} = \operatornamewithlimits{argmin}_{\bm{z} \in \R^n} \|\bm{A}\bm{M}\bm{z} - \bm{b}\|_2^2 ,
\end{align}
they produce iterates $\bm{z}_t$ that we transform to $\bm{x}_t = \bm{M}\bm{z}_t$. 
Note that if $\rank(\bm{A}\bm{M}) = \rank(\bm{A})$ and $\bm{\tilde{z}}$ solves this preconditioned problem, then $\bm{x} = \bm{M}\bm{\tilde{z}}$ solves \eqref{eq:LS}.

Our implementations of these solvers run until they hit an iteration limit, or until their solution reaches a termination criterion
\begin{equation}
\label{eq:main_stopping_criterion}
\|(\bm{A}\bm{M})^{\trans}\bm{r}\|_2/(\|\bm{A}\bm{M}\|_{\text{EF}} \|\bm{r}\|_2) \leq \rho ,
\end{equation}
where $\bm{r}=\bm{A}\bm{x}-\bm{b}$, $\|\bm{A}\bm{M}\|_{\text{EF}}$ is an \textit{estimate} of the Frobenius norm of $\bm{A}\bm{M}$, and $\rho$ indicates the user-specified error tolerance.
Appendix~\ref{sec:termination_criteria} provides more details about the design of the termination criteria and how $\|\bm{A}\bm{M}\|_{\text{EF}}$ is estimated.

\subsubsection{LSQR}

LSQR \cite{PS:1982} is an iterative method for approximately applying the pseudo-inverse of any matrix to a vector of conformable size.
It works by implicitly applying the Golub-Kahan bidiagonalization process to a 2-by-2 block matrix 
\firstrevnote{
$\begin{pmatrix}
  \bm{I}_{m \times m} &  \bm{A} \\ \bm{A^{\trans}} & \bm{0}_{n \times n}
\end{pmatrix}$.
}
The correctness of the recurrences used in this implicit bidiagonalization process is not obvious, and yet it is the key to LSQR's good numerical behavior in finite-precision arithmetic.

The behavior of LSQR in exact arithmetic is best understood through its interpretation as a Krylov subspace method.
Suppose for ease of exposition that we initialize it at the origin (i.e., we take $\bm{x}_0 = \bm{0}$ to obtain the normal equation residual $\bm{r}_0 = \bm{A}^{\trans}\bm{b}$) and we use \textit{no preconditioner}.
With these choices, LSQR's $t^{\text{th}}$ iterate $\bm{x}_t$ minimizes the loss function
\begin{equation}\label{eq:krylov_method_loss}
    L(\bm{x}) = \|\bm{A}(\bm{x} - \bm{A}^{\dagger}\bm{b})\|_2
\end{equation}
over the Krylov subspace
\begin{equation}\label{eq:krylov_subspace}
    K_t(\bm{A},\bm{b}) = \operatorname{span}\{\bm{r}_0, (\bm{A}^{\trans}\bm{A})\bm{r}_0,\ldots,(\bm{A}^{\trans}\bm{A})^{(t-1)}\bm{r}_0\}.
\end{equation}
This interpretation implies that LSQR is equivalent to the conjugate gradient method in exact arithmetic; see \cite{HS:1952:CGLS}.
Therefore, if we initialize LSQR at the origin, then its convergence guarantee takes the form
\begin{equation}\label{eq:lsqr_cg_converge}
    \|\bm{A}\left(\bm{x}_t - \bm{A}^\dagger\bm{b}\right)\|_2 \leq 2\left(\frac{\kappa - 1}{\kappa + 1}
    \right)^{t}\|\bm{A}\bm{A}^{\dagger}\bm{b}\|_2,
\end{equation}
where $\kappa = \operatorname{cond}(\bm{A})$ is the condition number of $\bm{A}$.
This error bound can be adapted to the case $\bm{x}_0 \neq \bm{0}$ simply by replacing $(\bm{x}_0,\bm{b}) \leftarrow (\bm{0},\bm{b} - \bm{A}\bm{x}_0)$.

Right-preconditioning $\bm{A}$ in LSQR does not affect the loss function $L$ in \eqref{eq:krylov_method_loss}.
However, it \textit{does} affect the initial residual $\bm{r}_0$ and more generally the Krylov subspace~\eqref{eq:krylov_subspace}; see~\cite{Bjorck:1996}.
This change in Krylov subspace buys us $\kappa = \operatorname{cond}(\bm{A}\bm{M})$ in~\eqref{eq:lsqr_cg_converge}.

\subsubsection{PGD}

Let $\bm{x}_{t} = \bm{M}\bm{z}_{t}$ be the algorithm's candidate solution at the end of iteration $t$, and let $\bm{r}_t = \bm{A}^{\trans}\left(\bm{b} - \bm{A}\bm{x}_t\right)$ be the corresponding residual vector.
The $t^{\text{th}}$ iteration of PGD ($t \geq 1$) has three steps:
\begin{enumerate}
    \item Compute $\Delta\bm{z} = \bm{M}^{\trans}\bm{r}_{t-1}$, which is the steepest-descent search direction for the right-preconditioned loss function $L(\bm{z}) = \|\bm{A}\bm{M}\bm{z} - \bm{b}\|_2^2$ at $\bm{z} = \bm{z}_{t-1}$.
    \item Using quantities computed in step 1, check the stopping criterion. If the criterion is satisfied we return $\bm{x}_{t-1}$.
    Otherwise, we we use a simple formula to compute $ \alpha_t = \operatornamewithlimits{argmin}\{L(\bm{z}_{t-1} + \alpha\Delta\bm{z}) \,:\, \alpha \geq 0\}$.
    \item Set $\bm{z}_{t} = \bm{z}_{t-1} + \alpha_t\Delta\bm{z}$ and $\bm{x}_{t} = \bm{M}\bm{z}_{t}$ and increment $t \leftarrow t + 1$.
\end{enumerate}
When initialized at the origin ($\bm{z}_0 = \bm{0}$), the convergence guarantee for PGD is
\begin{equation}\label{eq:steepest_descent_converge}
    \|\bm{A}\left(\bm{x}_t - \bm{A}^\dagger\bm{b}\right)\|_2 \leq 
    \left(\frac{\kappa^2 - 1}{\kappa^2 + 1}\right)^{t}
    \|\bm{A}\bm{A}^{\dagger}\bm{b}\|_2,
\end{equation}
where $\kappa = \operatorname{cond}(\bm{A}\bm{M})$~\cite{NW:2006:NumericalOptBook} [Theorem 3.3].
Note that this convergence rate is asymptotically worse than \eqref{eq:lsqr_cg_converge}.
However, if solutions of moderate accuracy suffice, then this preconditioned steepest-descent approach can be effective.

\section{The autotuning pipeline}
\label{sec:autotuning}

This section presents the details of our autotuning workflow introduced in Figure~\ref{fig:tuning_overview} and the tuning techniques for SAP-based randomized least squares methods. 

\subsection{Tuning description}
\label{sec:tuning_problem}

As shown in Figure~\ref{fig:tuning_overview}, the user needs to provide a \emph{tuning description} for the search space of tuning parameters and the tuning ``budget'' (we used the number of function evaluations, but total tuning time is a valid alternative).
Here, we explain the details of the tuning description.
We start with the terminology, and then we provide detailed explanations for the parameters.

In our context, \textit{task parameters} describe the given tuning problem (i.e., task), such as the problem size of the given overdetermined least squares problem.
\textit{Tuning parameters} are the parameters we want to optimize (i.e., SAP algorithm parameters), and \textit{objective values} are the values obtained after evaluating each configuration. 
\textit{Constant parameters} are user-provided, constant values for the tuning experiments and not tuned.
Parameters can have real, integer, or categorical types. Real variables have uncountably many values with a natural intrinsic ordering, integer variables have multiple values with an intrinsic ordering, and categorical variables have multiple categories without a natural ordering.
In our GP surrogate modeling, by default, all these variables are converted into real variables with a range of $[0,1]$.
As categorical variables have no intrinsic order, there are alternative methods such as a tree-based approach for categorical variables \mbox{\cite{luo2022hybrid}}.
In our transfer learning approach (Section~\mbox{\ref{sec:transfer_learning}}), we adopt a bandit (1-layer tree) algorithm to search    the categorical variable space, whereas we use GP to search the real and integer-valued variable space.

Table~\ref{tab:tuning_description} shows the list of information that the user needs to describe in order to run our autotuning pipeline. We give more details below.

\begin{table}[]
    \caption{User's tuning description (parameter space) for our autotuning pipeline for the SAP-based least squares. Our goal is to minimize \textit{wall\_clock\_time} subject to  $\textit{ARFE} \leq \textit{allowance\_factor}\times \textit{ARFE}_{\textit{ref}}$.
    The parameter search bounds and constant parameter values used in our experiments are provided in Table~\ref{tab:tuning_setup}.}
    \label{tab:tuning_description}
    \vspace{-1em}
    \footnotesize
    \centering
    \begin{tabular}{|m{2.4cm}|m{7.8cm}|m{1.5cm}|}
        \multicolumn{3}{l}{\textbf{Task parameters}} \\ \hline
        $m$ & Number of rows of the input matrix $\bm{A}$ & Integer \\ \hline
        $n$ & Number of columns of the input matrix $\bm{A}$ & Integer \\ \hline
        \multicolumn{3}{l}{\textbf{Tuning parameters}} \\ \hline
        \textit{SAP\_algorithm} & The SAP-based least squares algorithm (see Table~\ref{tab:sap_algorithms} for the SAP algorithms we consider in this paper) & Categorical \\ \hline
        \textit{sketching\_operator} & The sketching operator & Categorical \\ \hline
        \textit{sampling\_factor} & The size of the sketching matrix, $d = \textit{sampling\_factor} \times n$ & Real \\ \hline
        \textit{vec\_nnz} & Number of nonzero elements per row or column & Integer \\ \hline
        \textit{safety\_factor} & Parameter affecting the accuracy we request of LSQR/PGD & Integer \\ \hline
        \multicolumn{3}{l}{\textbf{Objective values}} \\ \hline
        \textit{wall\_clock\_time} & The execution time of the SAP algorithm & Real \\ \hline
        \textit{ARFE} & Residual error (approximate relative forward error) & Real \\ \hline
        \multicolumn{3}{l}{\textbf{Constant parameters}} \\ \hline
        \textit{num\_pilots} & Number of initial, random samples to evaluate  & Integer \\ \hline
        \textit{num\_repeats} & Number of repeats to run for each parameter configuration & Integer \\ \hline
        \textit{ref\_config} & Reference tuning parameter configuration that would lead to a sufficiently accurate residual error $\textit{ARFE}_{\textit{ref}}$ & List \\ \hline
        \textit{penalty\_factor} & Parameter to penalize inaccurate parameter configurations & Real \\ \hline
        \textit{allowance\_factor} & Parameter to determine an acceptable range of \textit{ARFE} & Real \\ \hline
    \end{tabular}
\end{table}

\subsubsection{Task and tuning parameters}

Task parameters consist of $m$ and $n$, representing the number of rows and columns of input matrix $\bm{A}$. These parameters indicate the given problem size.

Our tuning space consists of five parameters.
First, \textit{SAP\_algorithm} is a categorical parameter that determines which SAP algorithm to use.
Each option corresponds to a unique SAP-based randomized least squares algorithm (\textsf{QR-LSQR}, \textsf{SVD-LSQR}, or \textsf{SVD-PGD}) as described in Table~\ref{tab:sap_algorithms}.
This parameter is related to the two questions of the tuning opportunity trichotomy, how to generate the preconditioner (\textsf{TO2}) and what iterative solver to use (\textsf{TO3}).
The \textit{sketching\_operator} parameter is also a categorical parameter to choose the sketching operator.
Available options for \textit{sketching\_operator} in our work include SJLT and LessUniform.
Then, two ordinal parameters, \textit{sampling\_factor} (real type) and \textit{vec\_nnz} (integer type) are used to construct the sketching matrix (\textsf{TO1}).
For both SJLT and LessUniform, \textit{sampling\_factor} represents the size of $d$ relative to $n$ for the sketching operator, i.e., $d=sampling\_factor\times n$.
The \textit{vec\_nnz} parameter determines the sparsity of the sketching matrix.
With SJLT, \textit{vec\_nnz} refers to the number of non-zero entries per column (from~$1$ to~$d$), while with LessUniform, it refers to the number of non-zero entries per row (from~$1$ to~$m$).
Our fifth and final tuning parameter is \textit{safety\_factor}, an integer type.
This controls the requested error tolerance $\rho$ that the iterative solvers use in their stopping criterion \eqref{eq:main_stopping_criterion} (hence, this is related to \textsf{TO3}).
Specifically, we call solvers with error tolerance $\rho=10^{-(6+safety\_factor)}$.

The need for \textit{safety\_factor} as a tuning parameter is subtle.
It traces back to how the quality of a preconditioner affects not only an iterative solver's convergence rate, but also the metric that it uses in its termination criteria. 
Because of this fact it is important to check the accuracy of the algorithm's output against a reference value; if the accuracy is inadequate then the parameter configuration should be (heavily) penalized.
Tuning \textit{safety\_factor} lets us explore the potentially risky regions of parameter space for sketching operators without compromising our ability to compute solutions of a desired accuracy.

\subsubsection{Objective function}
\label{sec:objective_function}

The autotuning framework operates by evaluating and sequentially optimizing an objective function.
This function is central to the process as it quantifies the performance of a selected parameter configuration by running the SAP methodology and returning the objective values.

One obvious performance metric is \textit{wall-clock time}, which refers to the real time that elapses while the algorithm is running.
However, there is more to consider.
The accuracy of the solutions produced by the iterative \firstrevnote{methods} in the SAP algorithm can vary significantly with preconditioner quality and the \textit{safety\_factor} tuning parameter.
Therefore, our objective function takes into account both the wall-clock time and the returned solution's forward error.

Our choice of the forward error metric is motivated by the sensitivity of least squares solutions with respect to perturbations of $\bm{A}$.
Here it is well known that the sensitivity of $\bm{x}_*$ is proportional to $\kappa^2$ where $\kappa$ is the condition number of $\bm{A}$, while the sensitivity of $\bm{b} - \bm{A}\bm{x}_*$ is proportional to $\kappa$ (see, e.g., \cite[\S B.2]{RandLAPACK_book}).
This makes forward error with respect to $\bm{b} - \bm{A}\bm{x}$ a better metric when tuning for potentially ill-conditioned least squares problems.
We consider this metric in a relative sense where we normalize by $\|\bm{b} - \bm{A}\bm{x}\|$ for the approximate solution $\bm{x}$. 
This gives us the \textit{approximate relative forward error} (ARFE) 
\begin{equation}
\label{eq:arfe}
\textit{ARFE}=  
\frac{\|\bm{A}\bm{x}-\bm{A}\bm{x_{*}}\|_{2}}{\|\bm{A}\bm{x}-\bm{b}\|_{2}}.
\end{equation}
We emphasize that $\bm{x_{*}}$ represents the solution found by a direct least squares solver.
We also note that $\|\bm{A}\bm{x} - \bm{b}\|_2 \geq \|\bm{A}\bm{x}_* - \bm{b}\|_2$ is bounded away from zero for all problems we consider, because we have a sketch-and-solve ``presolve'' step (\mbox{Appendix~\ref{sec:sap_implementation}}) which would compute the exact solution without needing an iterative solver if we had $\bm{A}\bm{x}_* = \bm{b}$.
We use ${\textit{ARFE}}$ to determine the validity of the output of the SAP algorithms by comparing the results between direct and SAP methods.
If the $\textit{ARFE}$ of a parameter configuration exceeds a certain threshold, the configuration is considered a failure and its runtime should be penalized.

There are many ways that one might set the threshold for $\textit{ARFE}$ in a tuning application.
We propose the use of a ``reference'' parameter configuration, $\textit{ref\_config}$, to compute a reference \textit{ARFE} value, $\textit{ARFE}_{ref}$.
This reference configuration can be computationally expensive (e.g., high $\textit{sampling\_factor}$ and $\textit{vec\_nnz}$), since if $\bm{A}\bm{M}$ is close to an orthogonal projector then the solution's \textit{ARFE} should be close to the left-hand-side of the termination criterion \eqref{eq:main_stopping_criterion}.
The value $\textit{ARFE}_{ref}$ is used alongside another parameter called \textit{allowance\_factor}; if $\textit{ARFE} > \textit{allowance\_factor} \times \textit{ARFE}_{\textit{ref}}$ then we consider the parameter configuration to be~a~failure.

The remaining question is how to penalize such failing parameter configurations.
In our autotuning framework, we simply multiply the obtained wall-clock time with a user-given penalty value ($\textit{penalty\_factor}$).
In other words, for the objective function (which is used for surrogate modeling and parameter search), the return value is $\textit{penalty\_factor} \times \textit{wall\_clock\_time}$ if $\textit{ARFE} > \textit{allowance\_factor} \times \textit{ARFE}_{\textit{ref}}$.

\subsubsection{Constant parameters}
\label{sec:experiment_parameters}

The constant parameters are used to specify the values for the failure handling scheme discussed in Section~\ref{sec:objective_function} and for other constant inputs.
The $\textit{ref\_config}$, $\textit{base\_tol}$, $\textit{allowance\_factor}$, and $\textit{penalty\_factor}$, explained above, are categorized as constant parameters.
There are two more constant parameters, $\textit{num\_pilots}$ and $\textit{num\_repeats}$.
The $\textit{num\_pilot}$ parameter is the number of initial random samples.
In typical Bayesian tuning settings, when there is no available performance data, the user can start with a number of random samples.
The $\textit{num\_repeats}$ parameter is the number of runs for a parameter configuration for different random seeds.
This information is helpful because the randomized algorithms' performance can vary from run to run 
depending on the random seeds.
For our experiments in this paper, we take the average performance obtained from different random seeds.

\subsection{The autotuning pipeline}
\label{sec:autotuning_pipeline}

\begin{figure}[t]
    \centering
    \includegraphics[width=\textwidth]{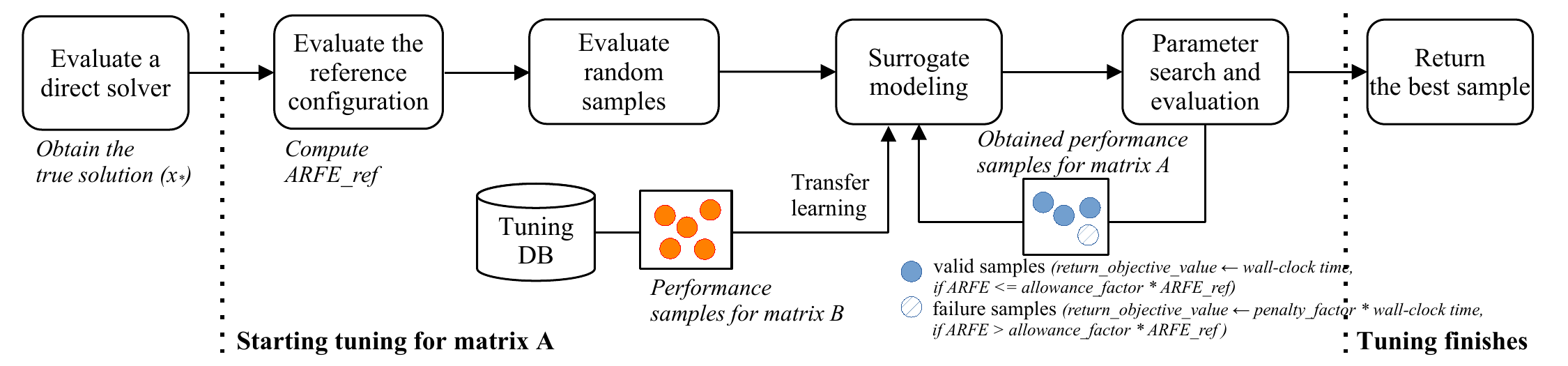}
    \vspace{-2em}
    \caption{Graphical presentation of the tuning procedure in the tuning pipeline.}
    \label{fig:tuning_pipeline}
\end{figure}

Figure~\ref{fig:tuning_pipeline} illustrates the procedure of the proposed tuning pipeline.
Before starting a function evaluation of the SAP algorithms, the tuner first runs and evaluates the given least squares problem using a direct solver.
The direct solver provides a solution $\bm{x_*}$ which will be used to compute $\textit{ARFE}$ of the output of each SAP algorithm.
For the first function evaluation of the SAP algorithms, we run the reference parameter configuration ($\textit{ref\_config}$) to compute the reference (accurate) residual error $\textit{ARFE}_{\textit{ref}}$.
As explained in Table~\ref{tab:tuning_description}, the reference configuration is a user-given constant parameter which is used to determine a sufficient level of accuracy desired from the tuned SAP algorithms.
The recommended practice is to choose a ``safe'' reference configuration
(e.g., as suggested by the theory), so that the SAP algorithm is guaranteed to achieve high accuracy, at the expense of much slower wall-clock time.

After the initial reference evaluation, the tuner follows one of two procedures, depending on whether or not we can rely on transfer learning from previously collected data. In the standard case (no transfer learning),  the tuner continues by evaluating a given number ($\textit{num\_pilots}$) of randomly chosen parameter configurations.
Then, the framework starts a Bayesian optimization-based tuning procedure using GP surrogate modeling.
Specifically, it iteratively builds a GP surrogate model and uses the model to select the next sample to evaluate (i.e., function evaluation).
For each function evaluation, we run the SAP algorithm using the suggested parameter configuration and measure the objective values: the wall-clock time and the residual error $\textit{ARFE}$.

\subsection{Transfer learning-based autotuning}
\label{sec:transfer_learning}

The optimal parameter configuration of the randomized least squares problem varies depending on the input matrix.
This fundamental challenge can make an autotuning approach prohibitive, because it requires the user to run autotuning individually for each different input matrix.
We advocate using transfer learning to overcome this challenge.
For a given input matrix to tune (which we call ``target task''), Transfer Learning-based Autotuning (TLA) aims to leverage previously collected performance samples obtained from different matrices (which we call ``source task(s)'').
If the source and target matrices have meaningful correlations, we can combine the surrogate models for the source and the target matrices for better parameter search on the current target task.
In other words, we can expect to achieve optimal tuning results with far fewer evaluations compared to tuning without transfer learning.

\begin{algorithm}
\caption{Procedure of TLA}
\label{alg:transfer_learning}
\begin{algorithmic}[1]
\STATE Evaluate the reference parameter configuration
\STATE Evaluate the historical best configuration from the source task(s)
\WHILE {there is remaining tuning budget}
  \STATE Choose a category ($\textit{SAP\_alg.}$ and $\textit{sketching\_operator}$) using UCB
  \STATE Choose the ordinal parameters ($sampling\_factor$, $vec\_nnz$, $safety\_factor$) using LCM
  \STATE Evaluate the suggested parameter configuration.
\ENDWHILE
\end{algorithmic}
\end{algorithm}

Algorithm~\ref{alg:transfer_learning} describes the procedure of TLA designed for tuning randomized algorithms.
As shown in Line 1, TLA also starts with evaluating the reference parameter configuration ($\textit{ref\_config}$), as the reference residual error $ARFE_{ref}$ is still needed to verify the accuracy of the samples for the target matrix.
In Line 2, we then evaluate the historical best configuration, which refers to the configuration that led to the best wall-clock time in the source matrices.
Note that at this point there is no information about the current target task at all, therefore, it is a natural assumption to start with the historical best sample from the source.
Lines 3--7 describe how we choose the parameter configuration for function evaluations from transfer learning for the given tuning budget.
For each function evaluation, we first choose the categorical variable, i.e., the SAP algorithm and the sketching operator, using the Upper Confidence Bound (UCB) bandit function (explained below).
Then we apply a GP-based multitask learning technique, called Linear Coregionalization Model (LCM), in order to learn from historical samples within the same chosen category from the source matrices.
Our two-step parameter search here is different from GPTune's default existing transfer learning which does not differentiate between the categorical variables and the ordinal parameters.
GPTune normalizes the parameter space to [0, 1].
It is known that GP would not perform well on categorical variables that have no natural intrinsic orders \mbox{\cite{luo2022hybrid}}.
We also observed that the GP-based transfer learning does not work well on categorical variables (we will compare this in Section~\ref{sec:experiments}).

More specifically, through the UCB bandit function \cite{auer2002using}, at each function evaluation we find the $\textit{SAP\_algorithm}$ and $\textit{sketching\_operator}$ that maximize the following:
\begin{equation*}
\label{eq:ucb}
    \operatorname*{argmax}_{\{\textit{SAP\_alg.},\textit{sketch.\_op.}\}} \left[ R_t(\{\small {\textit{SAP\_alg.},\textit{sketch.\_op.}\}} ) + c \sqrt { \frac{log\,t}{N_t( {\small \{\textit{SAP\_alg.},\textit{sketch.\_op.}\}} )} } \right] 
\end{equation*}
where $t$ is the number of samples (the number of source samples plus the number of target samples), $R_{t}$ is the reward function (which is the average performance of evaluations for the category), and $N_{t}(a)$ represents the number of samples for the category.
The constant parameter $c$ is used to balance between exploration and exploitation.
By default we set $c = 4$, but one could also use different values.
In the UCB function above, $R_t(\{\small {\textit{SAP\_alg.},\textit{sketch.\_op.}\}})$ returns high reward values for categories that have higher performance, and 
$\sqrt { \frac{log\,t}{N_t( {\small \{\textit{SAP\_alg.},\textit{sketch.\_op.}\}} )} }$ assigns more weights to less-explored categories.
Hence, a higher value of $c$ implies that we assign more weight to exploration.

Once we choose the category \{ $\textit{SAP\_algorithm}$, $\textit{sketching\_operator}$ \}, we use GP-based multitask learning, based on LCM, which constructs a joint GP model for multiple tasks~\cite{Liu:2021:PPoPP}.
For the $\delta$ given tasks (including the source and the target tasks), LCM builds a joint model of the functions that model the performance of the given tasks.
For an individual task $i$ among the given tasks, the modeled random function $f_{i}(x)$ is computed as follows:
$$ f_{i}(x) = \sum_{q=1}^{\delta}{a_{i,q}u_{q}(x)}$$
where $a_{i,q}$ are hyperparameters and each ${u_{q}}$ is an independent GP for the given task whose hyperparameters need to be learned.
For $u_{q}$, the covariance $k_{q}(u_{q}(x),u_{q}(x'))$ is based on a Gaussian kernel defined on a $\beta$-dimensional parameter space, as follows:
$$ k_{q}(x,x') = \sigma_{q}^{2} \exp\left(-\sum_{j=1}^{\beta}\frac{(x_{j}-{x_{j}}')^{2}}{I_{j}^{q}}\right) $$
where $\sigma_{q}$ and $I_{j}^{q}$ are hyperparameters that need to be learned for each dimension.
In short, we assume different lengthscale parameters $I^q_j$ for each dimension $j=1,\cdots,\beta$.
These parameters are learned in the GPTune surrogate modeling framework.
For more details about the mathematical model and how to train these hyperparameters, see \cite{GPTuneUserGuide2022}.

\subsection{Sensitivity analysis}

Sensitivity analysis can be used to study the effect of varying the parameters on the performance metric, with rigorous statistical uncertainty quantification.
In our context, sensitivity analysis refers to conducting an analysis of the tuning parameters to determine the ``sensitivity'' of the objective function to the individual parameters and their pairwise interactions. Such sensitivity analysis can provide useful insights into the tuning problem \cite{owen2014sobol}, which we illustrate using the so-called Sobol analysis technique.

We apply the Sobol analysis \cite{sobol2001global} to the surrogate model built using the given (historical) performance data samples.
The procedure of GPTune's sensitivity analysis is as follows.
GPTune first builds a surrogate model based on the provided performance data; then, using the surrogate model, it draws (many) performance data samples (relying on the Saltelli sampling sequence by default~\cite{saltelli2010variance}); and finally it performs a variance-based mathematical analysis.
The variance-based analysis computes the part of the total variance corresponding to each tuning parameter.
GPTune internally invokes SALib~\cite{Herman2017} for the Saltelli low-discrepancy sampling sequence (\cite{saltelli2010variance} as an extension of Sobol sampling sequence) and to compute the variance-based~analysis.

From the Sobol analysis output, we mainly consider two measures (Sobol indices): first-order indices (S1) and total effect (ST) indices.
For each tuning parameter, the first-order S1 index is the contribution to the output variance of the main effect of the tuning parameter. This measures the effect of varying each tuning parameter alone, but averaged over variations in other tuning parameters.
For each tuning parameter, the total effect index ST represents the total contribution (including interactions among parameters) of the tuning parameter to the response variance.
The value is obtained by summing all first-order and higher-order effects involving the given tuning parameter.
More mathematical details of these indices can be found in \cite{sobol2001global,saltelli2010variance}. 

\section{Experiments}
\label{sec:experiments}

In this section, we evaluate the proposed autotuning workflow using synthetically generated input matrices \firstrevnote{and three real-world datasets.}

\begin{table}[]
    \centering
    \caption{Properties of input matrices.
    \firstrevnote{Condition numbers are provided only to give an indication of how accurately matrix-vector products can be computed with each matrix in floating point arithmetic. Recall from Proposition \ref{prop:precond_quality} that the spectra of the \emph{preconditioned} versions of these matrices are independent of the spectra of the matrices themselves.}
    \revnote{SAP-based solvers would work equally well regardless of the condition numbers. Testing higher condition numbers will be important, if we are exploring the effects of numerical errors of the preconditioned matrix, but it is out of scope of this work.}
    \firstrevnote{In other words, we do not need to provide experiments for matrices with specific (high) condition numbers, as this would not affect the results.} 
    }
    \label{tab:input_matrices}
    \footnotesize
    \begin{tabular}{|m{2cm}|m{2cm}|m{5cm}|}
         \hline
         Matrix & Coherence & Condition number \\ \hline
         \firstrevnote{\textsf{GA}} & \firstrevnote{0.024} & \firstrevnote{3.349} \\ \hline
         \firstrevnote{\textsf{T5}} & \firstrevnote{0.638} & \firstrevnote{3.850} \\ \hline
         \firstrevnote{\textsf{T3}} & \firstrevnote{0.909} & \firstrevnote{6.795} \\ \hline
         \firstrevnote{\textsf{T1}} & \firstrevnote{1.0}   & \firstrevnote{2489.474} \\ \hline
    \end{tabular}
\end{table}

\subsection{Experimental setup}
\label{sec:experimental_setup}

We generate synthetic least squares problems, with matrices ($m=50{,}000, n=1{,}000$), following~\cite{Ma:2014:PMLR} and \cite{PW:2017:NewtSketch}. Our matrices are:

\begin{itemize}
    \item \textsf{GA}: Each row of the matrix $\bm{A}$ is generated from multivariate normal distribution, with covariance $\Sigma_{ij} = 2 \cdot 0.5^{|i-j|}$.
    \item \textsf{T5},\textsf{T3},\textsf{T1}: Each row of the matrix $\bm{A}$ is generated from multivariate $t$-distri\-bution with 5, 3, 1 degrees of freedom, with the covariance matrix above.
\end{itemize}
We then construct a vector $\bm{x}$ by filling it with $1$ for the first 10 entries and the last 10 entries, and $0.1$ for the remaining entries.
Then we obtain $\bm{b}$ from $\bm{b} = \bm{A}\bm{x} + \epsilon$, where $\epsilon \sim \mathcal{N}(0,(0.09)^2I_{n}))$.
This data is further described in Table~\ref{tab:input_matrices}.
The coherence of matrix $\bm{A}$ is computed as $\mu(\bm{A})=m\cdot\max_i\|\bm{U}_{(i)}\|_{2}^{2}$ for $i \in \{1 ... m\}$, where $\bm{U}_{(i)}$ is the i-th row of the matrix $\bm{U}$ which is an $m$ by $n$ matrix consisting of the $n$ left singular vectors of matrix $\bm{A}$ (Section 7.3.2 of \cite{MT:2020}).

Table~\ref{tab:tuning_setup} shows the search space of tuning parameters and the constant parameter values used in our experiments.
Note that we used $[1,10]$ for the $\textit{sampling\_factor}$ and $[1, 100]$ for the $\textit{vec\_nnz}$ parameter.
In principle, the sampling factor and the number of non zeros per rows/columns can have higher values.
However, it is known that setting these values too high can slow down the randomized algorithms to the point that they offer no advantage over traditional NLA algorithms.
For the reference parameter configuration ($\textit{ref\_config}$), we use \textsf{QR-LSQR} with an SJLT, while setting the sampling factor to 5 and $\textit{vec\_nnz}$ to 50 with a $\textit{safety\_factor}$ of 0.

We demonstrate the effectiveness of the surrogate-based autotuning approach by comparing with multiple tuning options: naive random search, exhaustive grid search, GP-based surrogate-based approaches (using GPTune), and a tree density estimator surrogate-based approach (using TPE), as we detail in the following.
\begin{itemize}
    \item Grid search: This runs evaluations for a prescribed grid of tuning parameter configurations, which will be detailed in Section~\ref{sec:grid_search}.
    \item Random search (\texttt{LHSMDU}): This performs non-surrogate random search using the Latin Hypercube Sampling with Multi-Dimensional Uniform (LHSMDU) method in~\cite{Moza:2020:Zenodo,DEUTSCH:2012}.
    \item Tree-Structured Parzen Estimator (\texttt{TPE})~\cite{bergstra2011algorithms}: This is another popular tree density estimator surrogate-based tuning approach, that captures the landscape using a density estimator instead of GP. We can use this approach for the surrogate modeling framework instead of GPTune. \firstrevnote{We used $\mathtt{hyperopt==0.2.5}$.}
    \item GP-based autotuning (\texttt{GPTune}): This is our proposed pipeline in Section~\ref{sec:autotuning} without using transfer learning.
    \item GPTune with transfer learning autotuning (\texttt{TLA}): 
    This is our proposed transfer learning-based autotuning discussed in Section~\ref{sec:transfer_learning}.
    We use some previously collected performance samples for a different, smaller matrix $\bm{B}$ ($m=10{,}000$, $n=1{,}000$) assuming that there are previously available performance data for~learning.
\end{itemize}
To cope with possible variations of tuning quality due to the stochastic nature of autotuning, we repeat each tuner with different seeds five times and report their mean and variance in the results.

The implementations of the SAP-based least squares methods are provided in PARLA~\cite{PARLA} (in the branch named ``newtonSketchLstsq'').
We use PARLA with a version of 0.1.5, where it internally uses \texttt{numpy} (1.21.2), \texttt{scipy} (1.7.3) and Intel MKL (2022).
For the hardware platform, we use an Intel E5-1680v2 (3GHz) computer system that contains 8 physical cores with 128GB of memory (1600MHz).
Note that the performance of the randomized least squares solvers (and most of other high-performance software) changes on different hardware platforms.
In other words, the user of the randomized least squares solvers needs to choose an optimal tuning parameter configuration to achieve a (near-)optimal performance.
Our autotuning framework is designed to help tune the parameter configurations automatically on a given hardware platform, while minimizing human efforts.

\begin{table}[]
    \caption{The parameter search bound and constant parameter values used for the experiments.}
    \label{tab:tuning_setup}
    \vspace{-1em}
    \footnotesize
    \centering
    \begin{tabular}{|m{2.4cm}|m{6cm}|m{1.5cm}|}
        \multicolumn{3}{l}{\textbf{Tuning parameters}} \\ \hline
        \textit{SAP\_algorithm} & \{ \textsf{QR-LSQR}, \textsf{SVD-LSQR}, \textsf{SVD-PGD} \} & Categorical \\ \hline
        \textit{sketching\_operator} & \{ SJLT, LessUniform \} & Categorical \\ \hline
        \textit{sampling\_factor} & [1, 10] & Real \\ \hline
        \textit{vec\_nnz} & [1, 100]  & Integer \\ \hline
        \textit{safety\_factor} & [0, 4] & Integer \\ \hline
        \multicolumn{3}{l}{\textbf{Constant parameters}} \\ \hline
        \textit{num\_pilots} & 10 & Integer \\ \hline
        \textit{num\_repeats} & 5 & Integer \\ \hline
        \textit{ref\_config} & [ \textsf{QR-LSQR}, SJLT, 5, 50, 0 ] & List \\ \hline
        \textit{penalty\_factor} & 2.0 & Real \\ \hline
        \textit{allowance\_factor} & 10.0 & Real \\ \hline
    \end{tabular}
\end{table}

\subsection{Grid search}
\label{sec:grid_search}

For the first set of experiments, we conduct a semi-exhaustive grid search%
\footnote{Our grid search uses a semi-exhaustive approach, where we evaluate a large number of parameter configurations to understand the performance surface on the search space, but it is infeasible to evaluate all the possible parameter configurations.}
to explore the ``true landscape'' of randomized least squares  solvers using the SAP paradigm.
The grid search evaluates parameter configurations for 10 different sampling factors ($\textit{sampling\_factor}\in \{ 1,2,3,\ldots,10 \}$)
and 19 different values of the number of non-zeros per column/row ($\textit{vec\_nnz} \in \{ 1,2,3,\dots,10,20,30,\dots,100 \}$) for the 6 categories (3 SAP algorithms and 2 sketching operators) and 3 safety factors (\{ 0, 2, 4 \}), i.e., $ 10 \times 19 \times 3 \times 2 \times 3 = 3,420 $ evaluations in total.
Naturally, such an exhaustive search is impractically expensive for automated parameter selection.
The purpose is to illustrate how tuning algorithmic parameters can affect the performance of solvers for different kinds of matrices.

\begin{figure*}
\centering
\includegraphics[width=\textwidth]{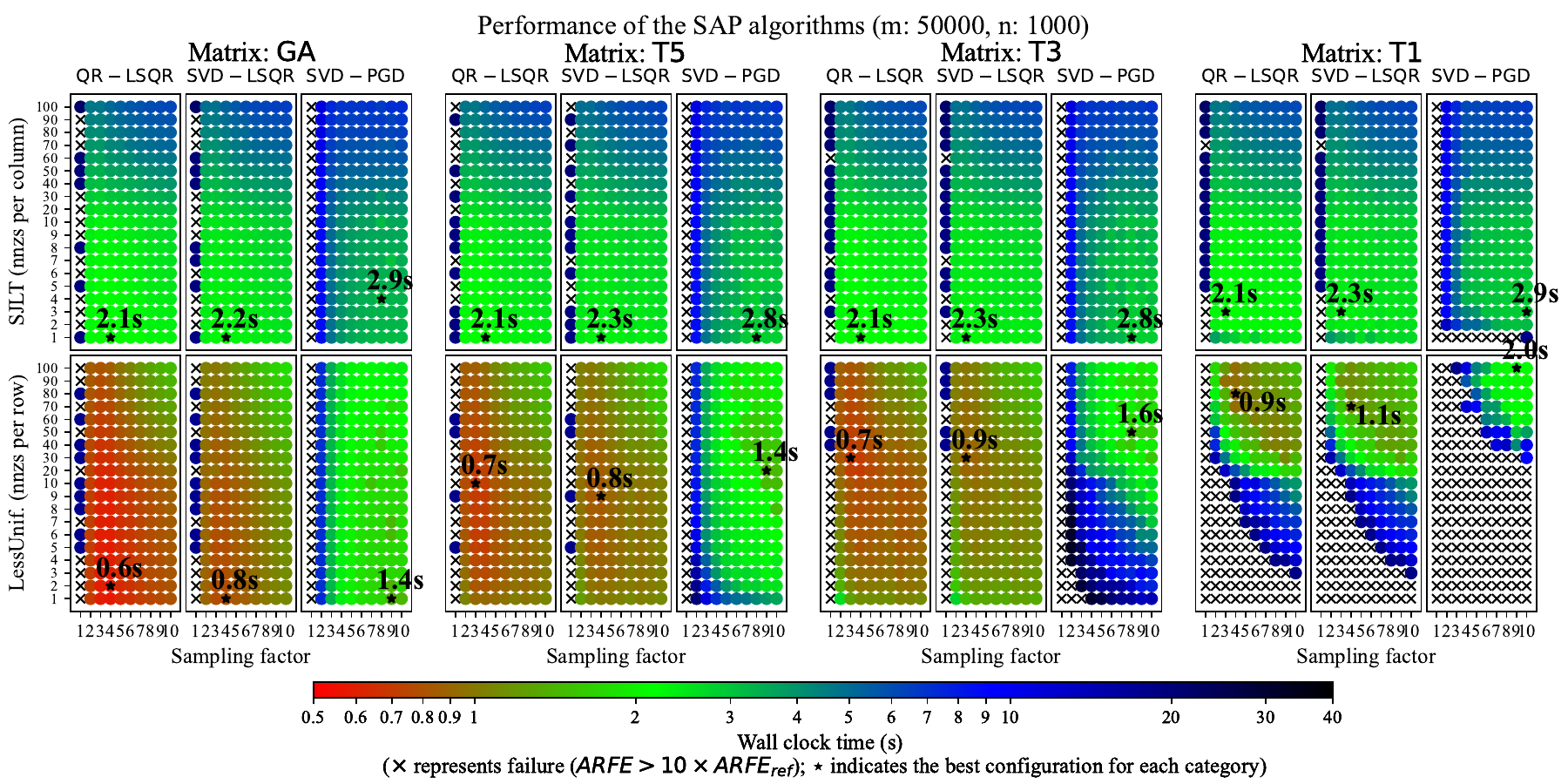}
\vspace{-2em}
\caption{\firstrevnote{Landscape of parameter configurations over a grid of combinations of parameters.} The labels on each plot represent the optimal performance and its parameter configuration in each category.
For each data point shown in the figure, we run three \texttt{safety\_factor} parameters ($0$, $2$, and $4$), and the plot shows the best result among these three \texttt{safety\_factor} parameter values.
}
\label{fig:grid_search}
\end{figure*}

Figure~\ref{fig:grid_search} shows the performance of different parameters on the four matrices, \textsf{GA}, \textsf{T5}, \textsf{T3}, and \textsf{T1}.
We observe that optimally-tuned LessUniform sketching leads to a significantly better wall-clock time than SJLT for all the three SAP algorithms.
This is partly because a $d \times m$ LessUniform operator has only $d \times vec\_nnz$ non-zeros, which is far fewer than the $m \times vec\_nnz$ non-zeros in an SJLT of the same size.
The data structures used in sketching also have a role here; applying a LessUniform operator lends itself to better cache efficiency than applying an SJLT when $\bm{A}$ and $\bm{M}$ are stored in row-major order (which is the standard for Python).

The three SAP algorithms yield varied wall-clock times.
We consistently see that \textsf{QR-LSQR} is faster than \textsf{SVD-LSQR} by a small margin, and also that \textsf{SVD-PGD} is materially slower and has many more $\textit{ARFE}$ failures (i.e., $\textit{ARFE}$ is higher than $10\times \textit{ARFE}_{\textit{ref}}$).
The limitations of \textsf{SVD-PGD} are particularly prominent for the high-coherence test matrix, \textsf{T1}.
The data seems to imply that PGD is far more sensitive to preconditioner quality than LSQR; this is consistent with theory, which states that LSQR still exhibits rapid convergence if only a few singular values of $\bm{A}\bm{M}$ are far from all others.
Of equal note is that using SJLTs instead of LessUniform dramatically reduces the number of \textit{ARFE} failures from \textsf{SVD-PGD}.
This is also consistent with theory, which states that SJLTs are suitable for data-oblivious sketching in ways that LessUniform operators are not.

Comparing the performance on different matrices with different characteristics, on the \textsf{GA} matrix with a low coherence, the best parameter configuration found by the grid search was using \textsf{QR-LSQR} and LessUniform with a sampling factor of 4 and only \firstrevnote{2} non-zero per row.
As coherence increases (e.g., \textsf{T5} and \textsf{T3} matrices), the optimal \textit{vec\_nnz} rises to \firstrevnote{10} and 30 respectively.
For the highly coherent \textsf{T1}, the best \textit{vec\_nnz} is \firstrevnote{80, given a sampling factor of 4}.
This shows that LessUniform requires significantly more non-zero entries for high-quality sketching as coherence increases -- a phenomenon that is not present with SJLTs.

This experiment demonstrates the necessity of performance tuning on SAP-based least squares.
On one hand, performing a grid search is obviously prohibitively expensive due to the large search space.
On the other hand, relying entirely on a ``safe'' parameter for a reliable accuracy can only achieve a sub-optimal performance result.
For example, compared to the reference configuration the optimal parameter configuration found by this grid, exhaustive search can achieve \firstrevnote{6.4x} (on \textsf{GA} matrix) and 3.9x (on \textsf{T1} matrix) faster wall-clock times.

\subsection{Autotuning results}
\label{sec:tuning_results}

\subsubsection{Comparison among the autotuners}
\label{sec:comparison_among_autotuners}

\begin{figure*}
\centering
\includegraphics[width=\textwidth]{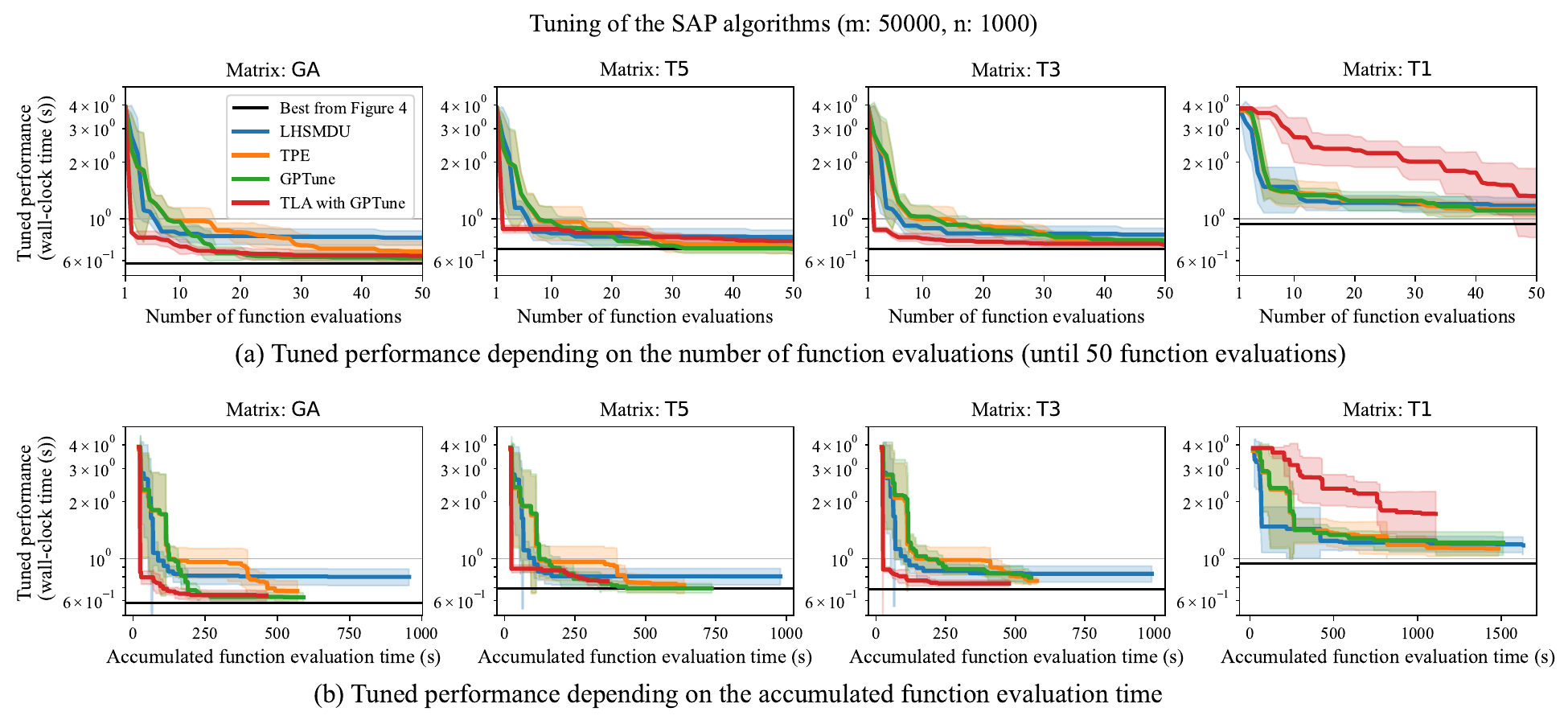}
\begin{adjustbox}{scale=0.65}
\begin{tabular}{c|c|c|c|c}
     Tuner & \textsf{GA} & \textsf{T5} & \textsf{T3} & \textsf{T1} \\ \hline
     LHSDMU          & \firstrevnote{1,017s} & \firstrevnote{1,035s} & \firstrevnote{1,076s} & \firstrevnote{1,894s} \\
     TPE             & \firstrevnote{685s}   & \firstrevnote{700s}   & \firstrevnote{756s}   & \firstrevnote{1,650s} \\
     GPTune          & \firstrevnote{788s}   & \firstrevnote{879s}   & \firstrevnote{759s}   & \firstrevnote{1,848s} \\
     TLA with GPTune & \firstrevnote{511s}   & \firstrevnote{549s}   & \firstrevnote{558s}   & \firstrevnote{1,286s} \\
\end{tabular}
\end{adjustbox}
\begin{adjustbox}{scale=0.65}
\centerline{(c) Accumulated function evaluation times for all the 50 function evaluations (average of five tuning runs for each tuner).}
\end{adjustbox}
\caption{\firstrevnote{Comparison of different tuners.} The first panel (a) compares the tuning results based on the number of function evaluations. The second panel (b) compares the tuning results based on the accumulated times for function evaluations (the summation of the runtimes of the evaluated parameter configurations; \firstrevnote{time is displayed until the shortest run among the five tuning runs for each tuner}), and the third panel (c) shows the accumulated function evaluation time of each tuner for the 50 function evaluations of the panel (a).
}
\label{fig:tuning_results}
\end{figure*}

We run and compare the autotuners discussed in Section~\ref{sec:experimental_setup} with far fewer evaluations (50 function evaluations), compared to the grid search experiment.
We show how the autotuning approach can attain (near-)optimal performance when there is a limited tuning budget.

Figure~\ref{fig:tuning_results} provides a comparison among the tuning results of the four different tuning options (\texttt{LHSMDU}, \texttt{TPE}, \texttt{GPTune}, and \texttt{TLA}), while comparing with the peak performance found by the (expensive) grid search from Section~\ref{sec:grid_search}.
In Figure~\ref{fig:tuning_results}, we observe that surrogate-based approaches (\texttt{TPE} and \texttt{GPTune}) tend to outperform \texttt{LHSMDU}, the random search method.
For instance, on the \textsf{GA} matrix, both \texttt{TPE} and \texttt{GPTune} yield better results than \texttt{LHSMDU}.
\firstrevnote{However, on the \textsf{T1} matrix, compared to \texttt{LHSMDU}, the surrogate-based approaches including \texttt{TPE}, \texttt{GPTune} and \texttt{TLA} do not offer a significant difference.}
Figure~\ref{fig:tuning_results} (b) illustrates the advantages of surrogate-based methods further by comparing tuning results against \emph{accumulated function evaluation times},
where we sum the runtimes taken by evaluating the parameter configurations.\footnote{In this paper, computational overhead of surrogate modeling and search is not considered, because we focus on reducing the cost of function evaluations. Modeling/search overhead depends on the parameter space and the number of data to train, not the application itself, i.e., if the size of the input matrix of the SAP solvers is large enough, the cost of surrogate modeling would be negligible. We can also balance between the overhead and the accuracy of the surrogate model, depending on the problem size (we leave it as our future work).}
Here, we observe that \texttt{LHSMDU} consumes significantly more function evaluation times, particularly on the \textsf{GA}, \textsf{T5}, and \textsf{T3} matrices, without converging to optimal performance.
\firstrevnote{In contrast, surrogate-based methods tend to focus on promising regions (as we obtain more parameter configurations), leading to optimized performance in a shorter time.
When we compare the surrogate modeling techniques, \texttt{GPTune} tends to perform better on \textsf{GA} and \textsf{T5}.}

\firstrevnote{The advantage of surrogate-based autotuning is more pronounced when transfer learning is used, except the case of \textsf{T1}.}
\texttt{TLA} uses source data from 100 function evaluations for random parameter configurations on a smaller $\bm{B}$ matrix (with $m=10,000$ and $n=1,000$), following the same matrix generation scheme,
among \textsf{T1}, \textsf{T3}, \textsf{T5}, and \textsf{GA}
, introduced in Section~\ref{sec:experimental_setup}.
In Figure~\ref{fig:tuning_results} (a), \texttt{TLA} attains near-optimal performance within a drastically reduced number of function evaluations.
In addition, Figure~\ref{fig:tuning_results} (b) illustrates that \texttt{TLA} completes evaluations much faster, indicating its propensity for evaluating promising parameter configurations, hence reducing the cost of computational resources.
This underscores the efficiency of transfer learning in obtaining suitable parameter configurations, especially when there is a limited computational~budget.

\firstrevnote{For a quantitative comparison, comparing the tuning result based on the number of evaluations (Figure~\mbox{\ref{fig:tuning_results}} (a)), on \textsf{GA}, the final tuned result of \texttt{LHSMDU} is 0.794s, which is about 1.37x slower than the optimal wall-clock time (0.580s) found by the expensive grid search in Section~\mbox{\ref{sec:grid_search}} and obtained after evaluating $44$ parameter configurations.
To obtain the same or better wall-clock time, \texttt{GPTune} and \texttt{TLA} need to evaluate only 27 parameter configurations and 14 parameter configurations, respectively.
This means that \texttt{GPTune} and \texttt{TLA} used 1.63x and 2.75x fewer parameter configurations than \texttt{LHSMDU}, respectively, to attain the same performance level.
On \textsf{T1}, on the other hand, the obtained tuning quality was not beneficial.
We suspect the reason is that \textsf{T1} is an extremely high coherence matrix and many parameter configurations fail in terms of the $\textit{ARFE}$ (refer to Figure~\ref{fig:grid_search}).
This opens a future research question about a failure handling mechanism considering the sensitivity of the penalty and allowance factors. We provide a more analysis in Appendix~\ref{sec:robustness}.
}

\subsubsection{On transfer learning}

\begin{figure}
    \centering
    \includegraphics[width=\textwidth]{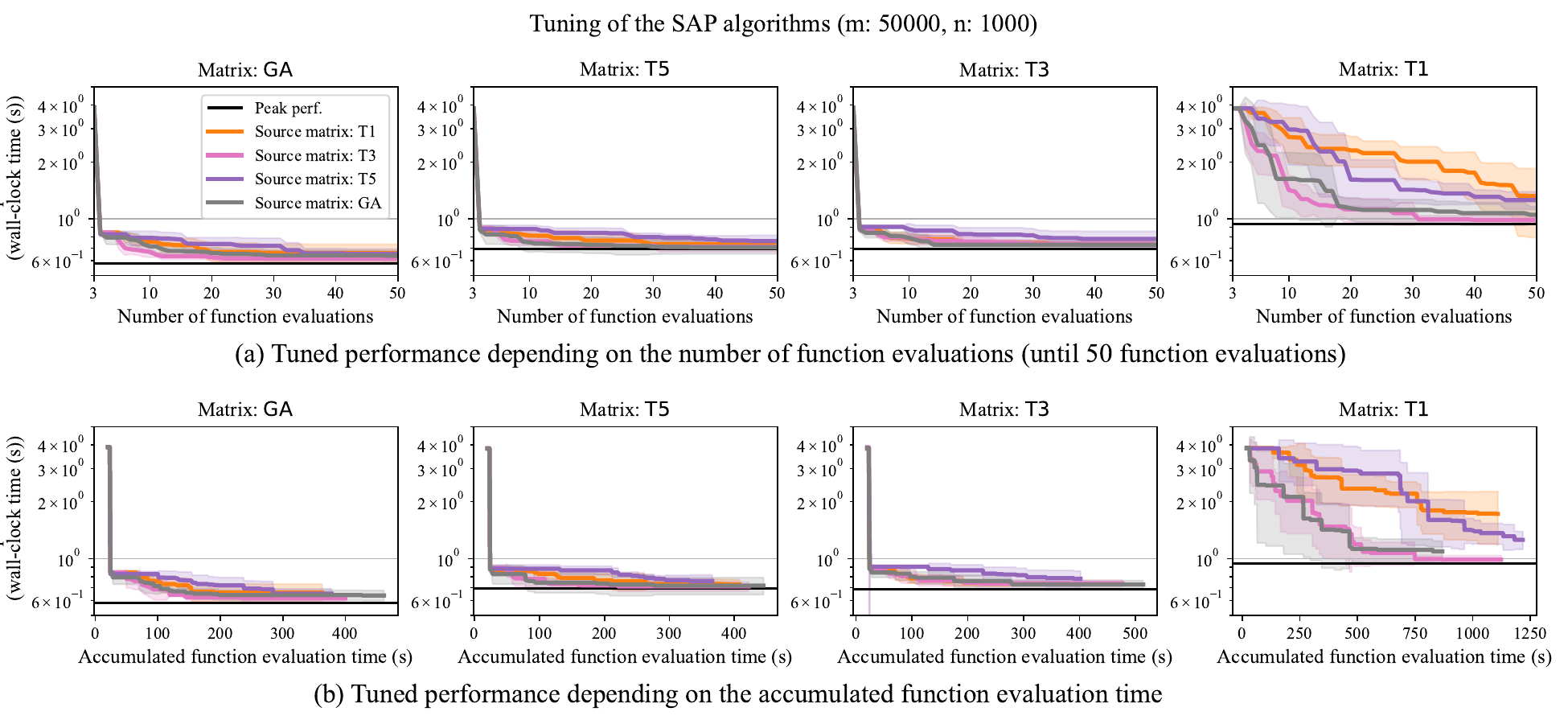}

\begin{adjustbox}{scale=0.65}
\begin{tabular}{c|c|c|c|c}
      & \textsf{GA} & \textsf{T5} & \textsf{T3} & \textsf{T1} \\ \hline
     Source matrix: \textsf{T1}   & \firstrevnote{430s} & \firstrevnote{506s} & \firstrevnote{550s} & \firstrevnote{1,286s} \\
     Source matrix: \textsf{T3}   & \firstrevnote{568s} & \firstrevnote{479s} & \firstrevnote{558s} & \firstrevnote{1,212s} \\
     Source matrix: \textsf{T5}   & \firstrevnote{487s} & \firstrevnote{549s} & \firstrevnote{586s} & \firstrevnote{1,308s} \\
     Source matrix: \textsf{GA}   & \firstrevnote{511s} & \firstrevnote{724s} & \firstrevnote{688s} & \firstrevnote{1,101s} \\
\end{tabular}
\end{adjustbox}
\begin{adjustbox}{scale=0.65}
\centerline{(c) Accumulated function evaluation times for all the 50 function evaluations (average of five tuning runs for each tuner).}
\end{adjustbox}
    \caption{\firstrevnote{Effect of choosing different source data for transfer learning.} For the X/Y axis labels of (a), (b), and (c), please refer to the caption of Figure~\ref{fig:tuning_results}.}
    \label{fig:TLA_source}
\end{figure}

In Figure~\ref{fig:TLA_source}, we further evaluate tuning quality of transfer learning (\texttt{TLA}) for varying source data with different characteristics, in order to see that choosing an appropriate source dataset would lead to a better tuning quality.
As discussed in Section~\ref{sec:experimental_setup} and \ref{sec:comparison_among_autotuners}, in Figure~\ref{fig:tuning_results} \texttt{TLA} takes a smaller matrix ($m=10{,}000$, $n=1{,}000$) with the same matrix generation scheme (among \textsf{T1}, \textsf{T3}, \textsf{T5}, and \textsf{GA}), as the source data.
In Figure~\ref{fig:TLA_source} we compare tuning quality when changing the source matrix.
The intuition is that using the same matrix generation scheme leads to similar matrix properties as we scale from the smaller to larger problems, which leads to the best tuning results.
\firstrevnote{In the experiments, we find that for the three matrices, \textsf{GA}, \textsf{T1}, and \textsf{T3}, \texttt{TLA} work reasonablly well regardless of the source data type.
On the \texttt{T1} matrix, it appears that the source data from the same \textsf{T1} was not effective. We suspect the reason is the small dataset was not effective to capture the behavior of many failing samples in the original dataset.
}

\begin{figure}
    \centering
    \includegraphics[width=\textwidth]{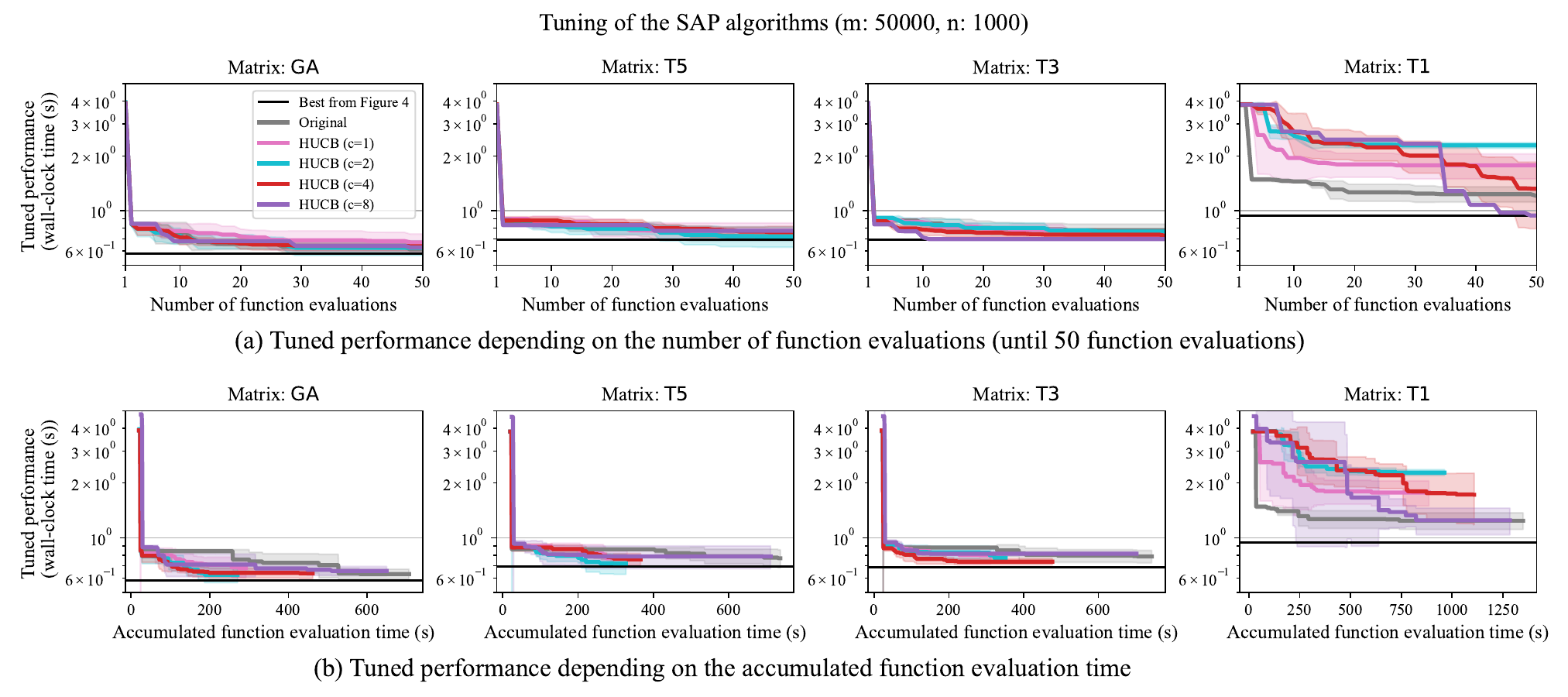}

\begin{adjustbox}{scale=0.65}
\begin{tabular}{c|c|c|c|c}
      & \textsf{GA} & \textsf{T5} & \textsf{T3} & \textsf{T1} \\ \hline
     Original   & \firstrevnote{825s} & \firstrevnote{928s} & \firstrevnote{850s} & \firstrevnote{1,590s} \\
     HUCB (c=1) & \firstrevnote{334s} & \firstrevnote{404s} & \firstrevnote{439s} & \firstrevnote{1,107s} \\
     HUCB (c=2) & \firstrevnote{336s} & \firstrevnote{340s} & \firstrevnote{539s} & \firstrevnote{1,045s} \\
     HUCB (c=4) & \firstrevnote{511s} & \firstrevnote{549s} & \firstrevnote{558s} & \firstrevnote{1,286s} \\
     HUCB (c=8) & \firstrevnote{708s} & \firstrevnote{895s} & \firstrevnote{908s} & \firstrevnote{1,590s} \\
\end{tabular}
\end{adjustbox}
\begin{adjustbox}{scale=0.65}
\centerline{(c) Accumulated function evaluation times for all the 50 function evaluations (average of five tuning runs for each tuner).}
\end{adjustbox}
    
    \caption{\firstrevnote{Effect of choosing different bandit constant parameter for transfer learning.} For the X/Y axis labels of (a), (b), and (c), please refer to the caption of Figure~\ref{fig:tuning_results}.
    Here, on (b), there are cases where one or more tuning runs finish 50 function evaluations with accumulated function evaluation time less than 400s; in this case, we plot the result until the shortest accumulated time among the five runs.
    }
    \label{fig:TLA_options}
\end{figure}

As discussed in Section~\ref{sec:transfer_learning}, for transfer learning (\texttt{TLA} in Figure~\ref{fig:tuning_results}) we use a mixture of the UCB bandit function and LCM-based multitask learning (which is GPTune's built-in technique~\cite{Liu:2021:PPoPP}, but it did not prove effective across different categorical parameters).
Figure~\ref{fig:TLA_options} compares the tuning quality of different transfer learning options.
Our \texttt{TLA} (equivalent to \texttt{HUCB ($c$=4)}) generally yields the best or close-to-best results, compared to using other constant parameter values for UCB (e.g., $c=1,2,8$) or GPTune's built-in LCM-based multitask learning (labelled as \texttt{Original}).
\firstrevnote{Our transfer learning method generally outperforms the LCM-based multitask learning, especially in terms of accumulated function evaluation time on \textsf{GA}, \textsf{T5}, and \textsf{T3}}.

\begin{figure*}
\centering
\includegraphics[width=\textwidth]{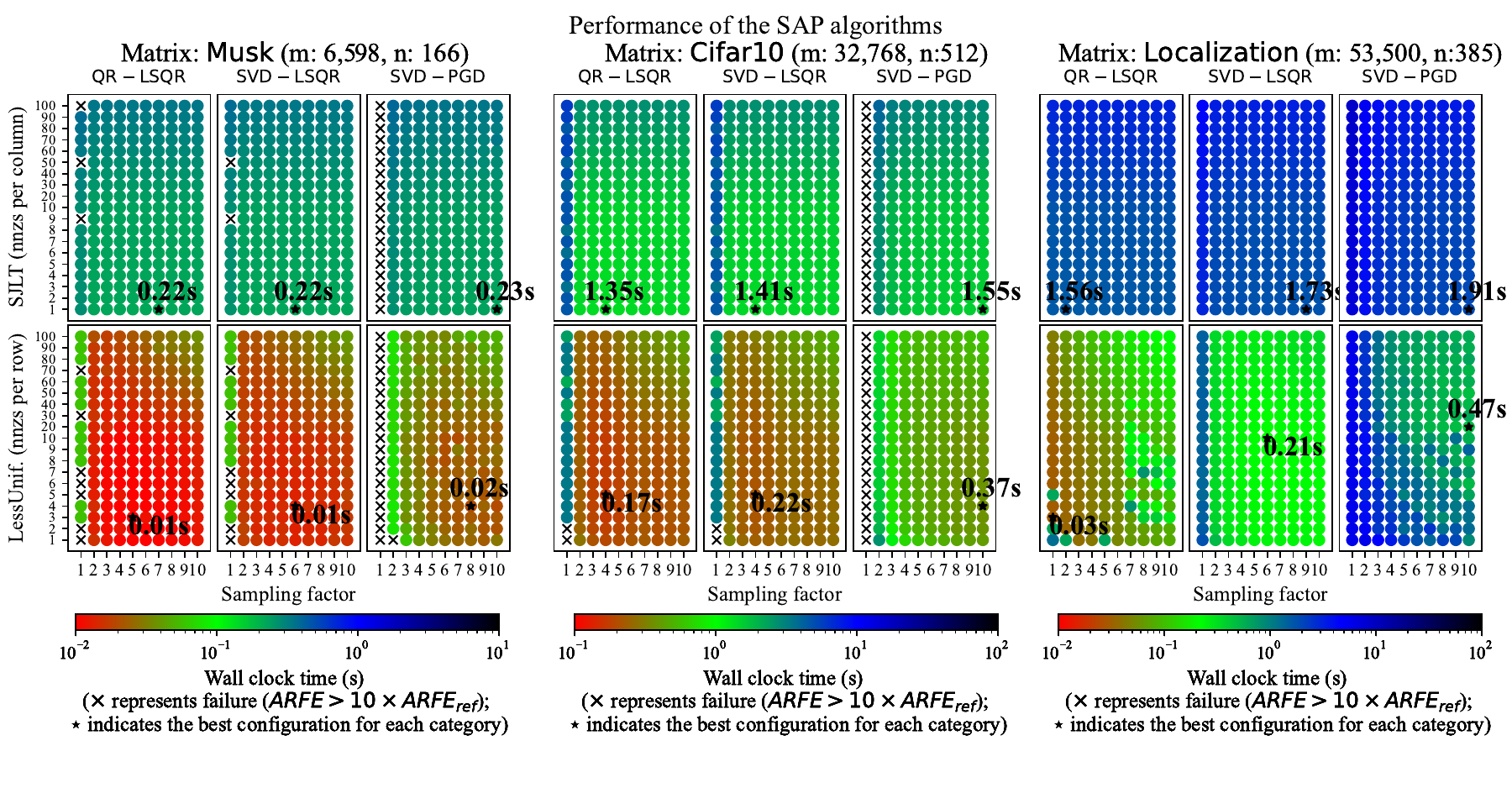}
\vspace{-3em}
\caption{\firstrevnote{Landscape of parameter configurations over a grid of combinations of parameters. The labels on each plot represent the optimal performance and its parameter configuration in each category.
For each data point shown in the figure, we run three \texttt{safety\_factor} parameters ($0$, $2$, and $4$), and the plot shows the best result among these three \texttt{safety\_factor} parameter values.}
}
\label{fig:grid_search_real_world}
\centering
\includegraphics[width=\textwidth]{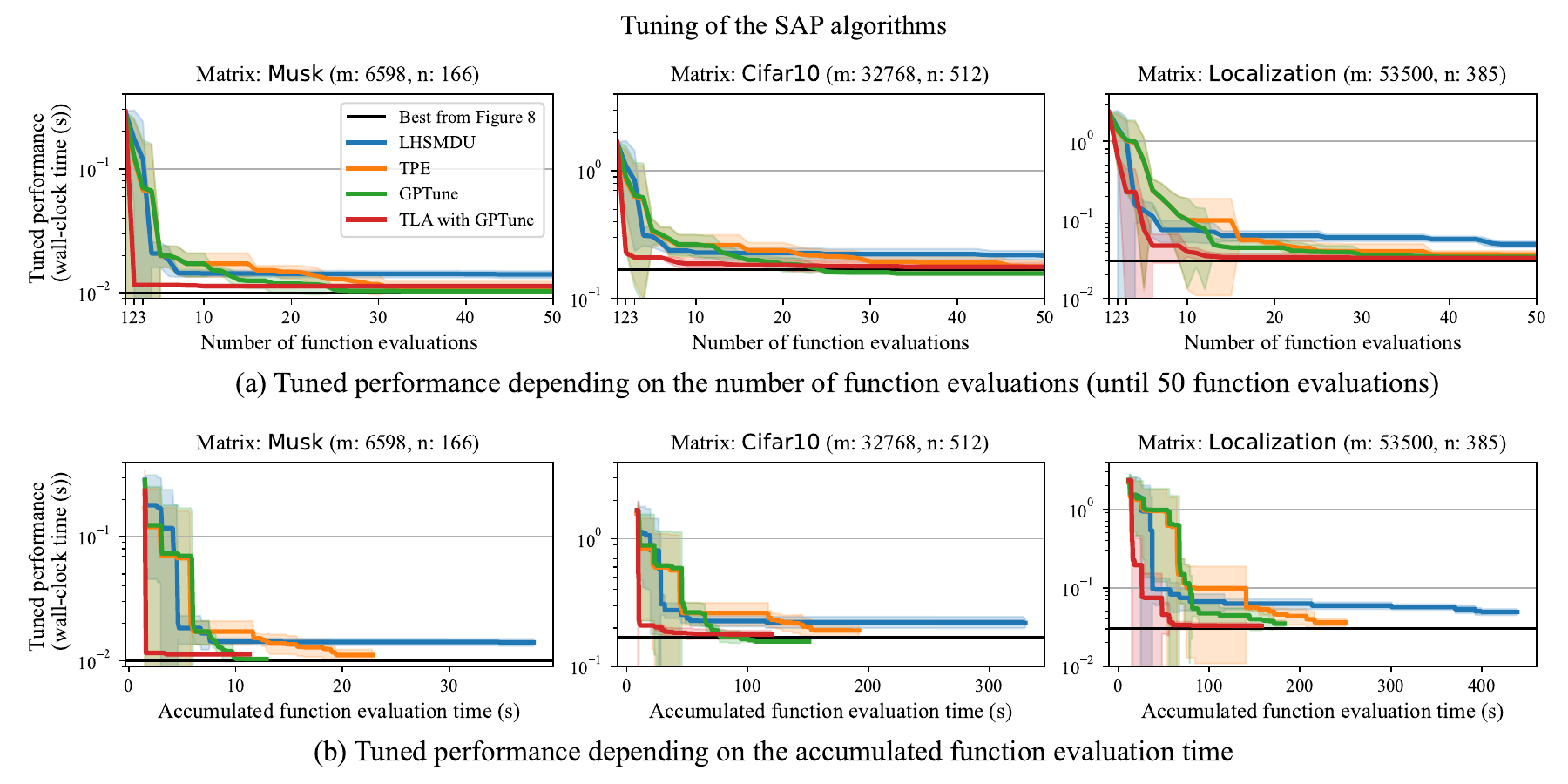}
\begin{adjustbox}{scale=0.65}
\begin{tabular}{c|c|c|c}
     Tuner           & \textsf{Musk} & \textsf{CIFAR-10} & \textsf{Localization} \\ \hline
     LHSDMU          & \firstrevnote{40s}     & \firstrevnote{347.1s}    & \firstrevnote{463.7s} \\
     TPE             & \firstrevnote{24.5s}   & \firstrevnote{216.2s}    & \firstrevnote{298.9s} \\
     GPTune          & \firstrevnote{14.8s}   & \firstrevnote{1974.0s}   & \firstrevnote{230.1s} \\
     TLA with GPTune & \firstrevnote{12.8s}   & \firstrevnote{1450.0s}   & \firstrevnote{183.2s} \\
\end{tabular}
\end{adjustbox}
\begin{adjustbox}{scale=0.65}
\centerline{(c) Accumulated function evaluation times for all the 50 function evaluations (average of five tuning runs for each tuner).}
\end{adjustbox}
\caption{\firstrevnote{Comparison of different tuners. For the X/Y axis labels of (a), (b), and (c), please refer to the caption of Figure~\ref{fig:tuning_results}.}
}
\label{fig:tuning_results_real_world}
\end{figure*}
\firstrevnote{

\subsection{Applicability of our tuning pipeline for real-world datasets}
\label{sec:experiments_real_world}

Using the synthetic matrices we have shown that our surrogate-based autotuning approach generally outperforms the random search such as \texttt{LHSMDU}.
In this section, we discuss experimental results of our autotuning approach for three real-world datasets to further validate the applicability our autotuning pipeline.

For the experiments, we chose three real-world problems, the \textsf{Musk} dataset, the \textsf{CIFAR10} dataset, and the \textsf{Localization} dataset.
The \textsf{Musk} dataset~\cite{misc_musk_(version_2)_75} has the input matrix size of $m=6,598$ and $n=166$, where the task is to classify/predict whether new molecules will be musks or not.
This is the smallest real-world datset we use in this paper.
\textsf{CIFAR-10}~\cite{cifar10} is an image classification dataset, and we used $m=32,768$ (number of instances) and $n=512$ (number of features), as the input matrix.
We then label the classes of CIFAR-10 into two groups following \cite{DLPM21_newtonless_TR}.
Lastly, \textsf{Localization}~\cite{misc_relative_location_of_ct_slices_on_axial_axis_206} is a regression dataset for Relative location of CT images, with a size of $m=53,500$ and $n=386$.

Figure~\ref{fig:grid_search_real_world} shows the grid search results to understand the true performance surface of the randomized least squares solvers.
Similar with the experiments in Figure~\ref{fig:grid_search} that uses the synthetic input matrices, we observe that ``LessUniform'' tends to perform better than SJLT, benefiting from the sparsity, and \textsf{QR-LSQR} usually outperforms the other two methods \textsf{SVD-LSQR} and \textsf{SVD-PGD}.
Regarding the sparsity, we can observe that these input data favor the use of a relatively low \textit{vec\_nnz}, compared to high-coherence synthetic matrices \textsf{T3} and \textsf{T1}.

In Figure~\ref{fig:tuning_results_real_world}, we conduct the same autotuning results with Figure~\ref{fig:tuning_results}, where we compare the autotuning qualities of \texttt{LHSMDU}, \texttt{TPE}, \texttt{GPTune}, \texttt{GPTune}, and \texttt{TLA}.
For the \texttt{TLA} experiment, we pre-collect 100 function evaluations for smaller matrices (we used random sampling); for the source matrices we used $m=2,048$ for \textsf{Musk}, $m=8,192$ for \textsf{CIFAR-10}, and $m=10,000$ for \textsf{Localization}.
Looking at Figure~\ref{fig:tuning_results_real_world} (b) we can observe that Bayesian optimization-tuning, both \texttt{TPE} and \texttt{GPTune}, is clearly more beneficial and effective than the simpler \texttt{LHSMDU}.
\texttt{TLA} is able to further improve the tuning quality by a significant margin, by exploiting previously collected data for smaller problems.

Note that \texttt{TLA} is to leverage previously collected data for different input problems, which can be down-sampled problem to approximate the larger problem.
For the experiment of \texttt{TLA} in Figure~\ref{fig:tuning_results_real_world}, for each real-world dataset, we used with different sizes of the data, \texttt{Musk} m: 2048, n: 166; \texttt{CIFAR-10} m: 8192, n:512; \texttt{Localization} m: 10000, n:385, where we collect $100$ samples (while this is a large number, collecting each sample for these smaller matrices is cheaper).

For example, on \textsf{Localization} that has the largest $m$ among the datasets, the final tuned result of \texttt{LHSMDU} is 0.049s, which is about 1.63x slower than the best result from the expensive grid search in Figure~\ref{fig:grid_search_real_world}.
To obtain the same or better wall-clock time, \texttt{TPE} and \texttt{GPTune} need 21 parameter configurations and 13 parameter configurations, respectively.
\texttt{TLA} needed only 6 parameter configurations showing the effectiveness of transfer learning.
In terms of the accumulated function evaluation time, to attain the 0.049s result, \texttt{LHSMDU} needs $399.5$s, \texttt{TPE} needs $171.5$s, and \texttt{GPTune} uses $88.6s$, showing that \texttt{GPTune} was the best among the non-transfer learning approaches.
Finally, \texttt{TLA} used only $48$s that is $8.3$x improvement compared to \texttt{LHSMDU}.

While this experiment demonstrates the generality and the applicability of our autotuning pipeline, one might have a question about the robustness of the framework with the choice of the allowance and penalty factor handling low accurate samples (i.e.,  $\textit{ARFE}$). Appendix~\ref{sec:robustness} shows some more experiments with different choices of the penalty and allowance factors.
}

\subsection{Sensitivity analysis}
\label{sec:sensitivityAnalysis}

Table~\ref{tab:sa_example} presents the sensitivity analysis results for the given tuning space described in Table~\ref{tab:tuning_setup} applied to the four synthetic matrices discussed in Section~\ref{sec:experimental_setup}.
The analysis quantifies the sensitivity of the five tuning parameters, \textit{SAP\_alg}, \textit{sketching\_operator}, \textit{sampling\_factor}, \textit{vec\_nnz}, and \textit{safety\_factor}.

Overall, higher Sobol index values (i.e., high S1 and ST values) indicate more sensitive parameters.
For the matrices considered, our analysis reveals two highly sensitive parameters, \textit{SAP\_alg} and \textit{sampling\_factor}, which determine the SAP algorithm and the size of the sketching matrix, respectively.
\textit{Sketching\_operator} also exhibits moderately significant sensitivity, and \textit{vec\_nnz} has the lowest sensitivity.
This observation resonates with our prior experiments (see Figure~\ref{fig:grid_search} and Figure~\ref{fig:grid_search_real_world}) that exhibited superior performance for specific SAP algorithm and sketching operator combinations, i.e., \textsf{QR-LSQR} and LessUniform, while the number of non-zeros showed moderate variation.
Even though ``number of non-zeros'' appears to be less sensitive, it should not be dismissed as unimportant. Its parameter space was set between 1 and 100, yet higher values could lead to different analysis results.
The \textit{safety\_factor} parameter, that controls the user-specified error tolerance of the SAP iterative \firstrevnote{method}, exhibits a high sensitivity score only for the \textsf{T1} matrix.
That is because error tolerance has a high influence in the case of the high-coherence \textsf{T1} matrix; 
higher tolerance can lead to higher error (which is regarded as a failure in our analysis), especially on high-coherence matrices.

To further interpret the sensitivity analysis results in Table~\mbox{\ref{tab:sa_example}}, we examine more cases.
When the index value is \firstrevnote{approximately zero,  because the interval encompasses zero with a width greater than the absolute index value} (e.g., S1 of \textit{sketching\_operator} on \textsf{GA}), it indicates a potential negligible effect of the input variable on the output, albeit with substantial uncertainty.
The wide interval containing zero undermines the conclusiveness of the variable's insignificance. Even if with small uncertainty, the absolute value of an index close to zero indicates a negligible effect compared to other variables.
If the index value is distant from zero, two subcases emerge:
(a) If the interval width is greater than the absolute index values but does not contain zero (e.g., \firstrevnote{S1 of  \textit{SAP\_alg} on \textsf{CIFAR-10}}), it signifies a significant effect with considerable uncertainty in quantifying the influence.
(b) If the interval width is smaller than the absolute index values and does not contain zero (e.g., \firstrevnote{ST of \textit{sketch\_operator} on \textsf{Localization}}), it denotes a significant effect with high confidence.

The advantage of conducting sensitivity analysis, particularly using surrogate-based techniques like Sobol \cite{owen2014sobol}, is its ability to afford an approximate understanding of the parameters.
This method is particularly useful when users operate under limited tuning budgets, in terms of either the number of evaluations or computational time. Depending on this budget, users might prefer to reduce the number of tuning parameters.
For example, if the tuning budget is very limited, they might decide to only adjust the ``sampling factor'' and ``sketching operator,'' keeping the randomized algorithm and sketching operator type constant. This form of sensitivity analysis, facilitated by surrogate-based approaches, is especially beneficial within the field of NLA, where such tuning exercises can significantly influence performance.
\firstrevnote{The sensitivity analysis has limitations, for example, it would require many samples for the analysis. However, we provide the results as supplementary results and additional analysis, rather than as the part of the required tuning process.}

\begin{table}[]
    \footnotesize
    \caption{
    \firstrevnote{Sobol sensitivity analysis results for the three input matrix \texttt{Musk}, \texttt{CIFAR-10} and \texttt{Localization}, with different sizes (\texttt{Musk} m: 2048, n: 166; \texttt{CIFAR-10} m: 8192, n:512; \texttt{Localization} m: 10000, n:385). The table shows the S1 and ST values with their confidence interval for each tuning parameter.}
    To run the Sobol analysis, the surrogate model is built using 100 function evaluation samples obtained from random sampling~\cite{DEUTSCH:2012} and the Sobol analysis draws 512 Saltelli sample sequence from the surrogate model for analysis.}
    \centering
    \footnotesize
    \begin{tabular}{c|c|c|c|c|c}
          \multicolumn{6}{c}{S1 (S1\_conf)} \\ \hline
          Input matrix & SAP\_alg & sketch\_operator & sampling\_factor & vec\_nnz & safety\_factor \\ \hline \hline
          \firstrevnote{Musk}          & \firstrevnote{0.0 (0.04)}  & \firstrevnote{0.69 (0.11)} & \firstrevnote{0.07 (0.1)}  & \firstrevnote{0.03 (0.02)} & \firstrevnote{0.0 (0.01)} \\
          \firstrevnote{CIFAR-10}      & \firstrevnote{0.02 (0.06)} & \firstrevnote{0.11 (0.06)} & \firstrevnote{0.32 (0.25)} & \firstrevnote{0.01 (0.01)} & \firstrevnote{0.01 (0.02)} \\
          \firstrevnote{Localization}  & \firstrevnote{0.1 (0.06)}  & \firstrevnote{0.59 (0.08)} & \firstrevnote{0.11 (0.09)} & \firstrevnote{0.04 (0.03)} & \firstrevnote{0.0 (0.01)} \\
    \end{tabular}
    \begin{tabular}{c|c|c|c|c|c}
          \multicolumn{6}{c}{ST (ST\_conf)} \\ \hline
          Input matrix & SAP\_alg & sketch\_operator & sampling\_factor & vec\_nnz & safety\_factor \\ \hline \hline
          \firstrevnote{Musk}         & \firstrevnote{0.16 (0.07)} & \firstrevnote{0.69 (0.12)} & \firstrevnote{0.23 (0.08)} & \firstrevnote{0.03 (0.01)} & \firstrevnote{0.01 (0.0)} \\
          \firstrevnote{CIFAR-10}     & \firstrevnote{0.48 (0.26)} & \firstrevnote{0.18 (0.09)} & \firstrevnote{0.8 (0.33)}  & \firstrevnote{0.01 (0.01)} & \firstrevnote{0.03 (0.02)} \\
          \firstrevnote{Localization} & \firstrevnote{0.27 (0.08)} & \firstrevnote{0.6 (0.08)}  & \firstrevnote{0.25 (0.07)} & \firstrevnote{0.04 (0.01)} & \firstrevnote{0.0 (0.0)} \\
    \end{tabular}
    \label{tab:sa_example}
\end{table}

\section{Related work}
\label{sec:related_work}

The use of empirical \textit{semi-exhaustive autotuning} methods in NLA is by now well-established \cite{montero2014beginner}. 
In the absence of analytic algorithm performance models like roofline analysis \cite{williams2009roofline, lo2014roofline} that take all the tuning parameters into account, the NLA community has adopted autotuning techniques for deterministic algorithms. 
Early work includes 
PHiPAC \cite{PHiPAC}, which provided a methodology for developing portable high-performance linear algebra libraries in ANSI C, and 
ATLAS \cite{whaley1998automatically}, which automated the tuning process for BLAS \cite{BLAS3}. 
Subsequent work explored parameter search for algorithmic parameters \cite{bernabe2014tuning} and online autotuning for sparse matrix computational kernels \cite{vuduc2005oski}. 
More recently, BLIS \cite{BLIS} has rapidly instantiated high-performance dense linear libraries. 
Our approach differs from these works, as our approach uses a surrogate model-based Bayesian optimization, using Gaussian process regression.
Our surrogate-based approach does not require any prior information about the application's performance.

The idea of surrogate-based autotuning has been applied to a variety of domains, including machine learning services \cite{bernabe2014tuning, Falkner:2018, Golovin:2017:KDD, liao2023efficient} and high-performance computing applications \cite{Liu:2021:PPoPP, Menon_2020_IPDPS, Roy:2021:PLDI, luo_nonsmooth_2021, luo2022hybrid}.
To the best of our knowledge, our work is the first in adopting a surrogate-based autotuning approach in the context of RandNLA, while providing a detailed empirical analysis for SAP-based least squares methods.

\section{Conclusions \revnote{and future work}}
\label{sec:conclusion}

In this paper, we presented a surrogate model-based autotuning approach for understanding and optimizing performance of RandNLA algorithms.
In particular, we described how we designed a tuning pipeline to address RandNLA-specific tuning challenges and presented the benefit of surrogate-based autotuning using the randomized least squares problem.
In our evaluation, we provided a detailed performance analysis of SAP-based randomized least squares and demonstrated that the surrogate-based tuning approach (using transfer learning) can attain near optimal performance faster than simple tuning approaches such as grid and random search.
The tuning framework and experiment scripts used in this paper are provided at \mbox{\url{https://github.com/gptune/parla-tuning}}.

\revnote{
To the best of our knowledge, our work is the first attempt to use a surrogate-based empirical autotuning pipeline in the context of RandNLA algorithms, and this opens up many new future research directions as well as the challenges to be investigated for achieving higher tuning quality.
For future work, first, we can explore a larger tuning space and/or a higher dimensional space, to further validate the efficacy of the surrogate-based autotuning approach.
The tuning space used in our experiments was made with a reasonable reduction of the space by the human insight, but it would be valuable if the tuner can work on a larger space (e.g., more sketching operators and/or preconditioner options) for better usability.
Second, more investigation is needed for more robust error handling methods for strongly constrained situations, where the user aims to tune the solver to achieve highly accurate solutions, and therefore there are many invalid configurations in terms of the residual accuracy.
As we have observed from our experiments (Appendix~\ref{sec:robustness}), while we presented an error handling scheme that uses $\textit{ARFE}$ and the penalty/allowance factors (see Section~\ref{sec:autotuning_pipeline}), it turns out that the scheme is simple and not always effective in transfer learning, especially on strongly constrained situations.
We expect that more sophisticated error handling schemes can resolve such an issue, but leave it as future work.
Third, the sensitivity analysis-based parameter reduction technique can be further researched.
One can try more sophisticated sensitivity analysis techniques that are more effective to reduce the tuning space of RandNLA problems.
Lastly, more extensive experiments can be done with more real-world, larger-scale matrices.
As discussed in Section~\ref{sec:envisioned_use_cases}, one practical use case of the surrogate-based autotuning approach is to tune the parameters on a smaller (down-sampled) data matrix and then use the tuned parameter to solve a larger matrix or equation.
We also leave it as future work.
}

\section*{Acknowledgement}

We thank Yang Liu, Osman A. Malik and Xiaoye S. Li for discussions and comments on this paper.
Haoyun Li was at University of California, Berkeley, when he worked on this project. He is now with Georgia Institute of Technology.

\appendix

\section{Our implementations}\label{sec:sap_implementation}
Here, we provide the implementation details of the
SAP algorithms in Table~\ref{tab:sap_algorithms} and explain the difference from the work on which each algorithm is based.
Each of these SAP algorithms uses the decomposition of $\bm{S}\bm{A}$ computed during preconditioner generation to obtain
\[
\bm{z}_{\mathrm{sk}} = \operatornamewithlimits{argmin}_{\bm{z}}\|\bm{S}\left(\bm{A}\bm{M}\bm{z} - \bm{b}\right)\|_2^2 ,
\]
which takes $O(dn)$ added time given $\bm{S}\bm{b}$.\footnote{For example, if $\bm{Q}$ is the orthogonal factor from QR of $\bm{\hat{A}}$, then we have $\bm{z}_{\text{sk}} = \bm{Q}^{\trans}\bm{S}\bm{b}$.}
In what can be interpreted as a presolve step, the iterative solver is then initialized at either $\bm{z}_{\text{sk}}$ (if $\|\bm{A}\bm{M}\bm{z}_{\text{sk}} - \bm{b}\|_2^2 < \|\bm{b}\|_2^2$) or the zero vector.

\subsection{QR-LSQR and Blendenpik}
Blendenpik uses the same preconditioned LSQR approach as the algorithm dubbed ``QR-LSQR'' in \cref{tab:sap_algorithms}, but it did not use our presolve step.
Blendenpik also differs from our algorithm in that it falls back on LAPACK if the factor $\bm{R}$ is nearly-singular.
Finally, we note that while we use sparse sketching operators, \cite{AMT:2010:Blendenpik} used subsampled randomized fast trigonometric transforms.

\subsection{SVD-LSQR and LSRN}
LSRN~\cite{MSM:2014:LSRN} originally assumed the sketching operator $\bm{S}$ was a matrix of iid mean-zero Gaussian random variables of appropriate variance.
Such sketching operators are impractical in our setting of solving dense least squares problems on shared-memory machines, since computing $\bm{S}\bm{A}$ would be extremely expensive.
That the original LSRN implementation used Gaussian sketching operators was important for their setup.
That made it possible to predict the extreme singular values of $\bm{A}\bm{M}$ accurately, without making assumptions on $\bm{A}$; and having such estimates opened up the possibility of using the Chebyshev semi-iterative method (see \cite{GV:1961}), which is more favorable than LSQR in distributed computation. 

\subsection{SVD-PGD and NewtonSketch}
\label{a:newtonsketch}

NewtonSketch, also known as Iterative Hessian Sketch, is a template algorithm developed by \cite{PW:2016:HessSketch,PW:2017:NewtSketch}.
It traditionally entails constructing a new sketching operator (and applying it to a new data matrix) in each iteration.
In that context, NewtonSketch is suitable for convex optimization problems which feature non-quadratic objective functions or problems with constraints other than linear~equations, as well as for optimization in distributed environments \cite{derezinski2020debiasing}.

NewtonSketch has a natural specialization for least squares which entails sampling and applying only one sketching operator.
This specialization can be viewed as an SAP approach where PGD is the iterative method.
We note that the slower convergence rate of PGD compared to (preconditioned) LSQR has motivated the use of alternative step sizes and so-called \textit{heavy ball momentum} for NewtonSketch \cite{OPA:2019,LP:2019}, as well extensions that incorporate sketching directly into the iterative method \cite{Der22_Stochastic_TR}.
Still, if solutions of high accuracy are required, then it is indeed advisable to use a ``traditional'' iterative method such as LSQR rather than PGD or a variant thereof~\cite{LP:2019}.
This leads to a potentially non-trivial tradeoff space for an autotuner.
The effect of the sparsity of a sketching operator on the convergence behavior of NewtonSketch was recently studied by \cite{DLPM21_newtonless_TR}, who provided theoretical analysis showing that extremely sparse sketches, such as the data-oblivious LessUniform, can lead to good performance.

\section{Termination criteria}
\label{sec:termination_criteria}
Iterative solvers accept accuracy parameters that are checked against termination criteria in each iteration.
The question of what the termination criteria should be is nontrivial.
The classical NLA literature addresses this problem through the concept of \textit{backward error}, which focuses on determining the extent of perturbation required in the problem data for the current solution to precisely solve the perturbed problem.

In its original form, LSQR uses two termination criteria, both concerned with bounding backward error.
One criterion is used for systems that appear to be consistent (which are typically square or underdetermined) and another is used for systems that appear to be inconsistent (which are typically overdetermined).
We say \textit{appear to be} in these conditions because LSQR actually checks both conditions at every iteration.

We have found that LSQR's termination criterion for consistent systems is easily triggered when solving to lower accuracy requirements, and that the quality of the returned solution is much worse than what one would expect.
Therefore, in our implementations we modified LSQR and implemented PGD so as to only use LSQR's termination criterion for inconsistent problems, namely stopping the algorithm when
\begin{equation*}
\|(\bm{A}\bm{M})^{\trans}\bm{r}\|_2/(\|\bm{A}\bm{M}\|_{\text{EF}} \|\bm{r}\|_2) \leq \rho
\end{equation*}
where $\bm{r}=\bm{A}\bm{x}-\bm{b}$, $\|\bm{A}\bm{M}\|_{\text{EF}}$ is an \textit{estimate} for the Frobenius norm of $\bm{A}\bm{M}$,\footnote{LSQR's estimate is updated at each iteration and is nondecreasing across iterations. PGD takes $\|\bm{A}\bm{M}\|_{\text{EF}} = \sqrt{n}$ for all iterations, which is reasonable in our context.} and $\rho$ indicates the user-specified error tolerance.
Note that even though we provide a certain value of $\rho$, the resulting residual accuracy from LSQR and PGD can vary depending on the parameter configuration; this leads us to regard $\rho$ as a tuning parameter and is discussed in Section~\ref{sec:autotuning}.

\firstrevnote{
\section{Effect of allowance and penalty factors}
\label{sec:robustness}
Our tuning pipeline accepts several user-given constant parameters (see Table~\ref{tab:tuning_setup}) to give a penalty to low accuracy samples (i.e., high $\textit{ARFE}$) in Bayesian optimization.
The choice of the constant parameters would affect the overall tuning quality.
While our default setting~\footnote{$\textit{allowance\_factor}=10.0$, $\textit{penalty\_factor}=2.0$ and the default parameter (QR-LSQR with SJLT, sampling factor of 5 and vec\_nnz of 50, with safety\_factor of 0 that determines $\textit{ARFE\_{ref}}$)} chosen by the authors is expected to perform generally well, one might want to test with other choices of the configurations.
Figure~\ref{fig:tuning_results_real_world_soft_strong} provides two more experiments with different settings, one is more strongly constrained setting in terms of the expected $\textit{ARFE}$ and the other is even less constrained $\textsf{ARFE}$ than the default setting.
We observe that for softly constrained setting (Figure~\ref{fig:tuning_results_real_world_soft_strong} (b)) we still achieve good tuning results.
However, for strongly constrained settings (a), where there can be many more failures in terms of $\textit{ARFE}$, the autotuning quality, especially non-TLA approaches such as \texttt{TPE} and \texttt{GPTune} suffers some lower quality.
A more sophisticated failure handling cost can help further improve the overall benefit of autotuning, but we would leave it as future work.

\begin{figure}
    \centering
    \includegraphics[width=\linewidth]{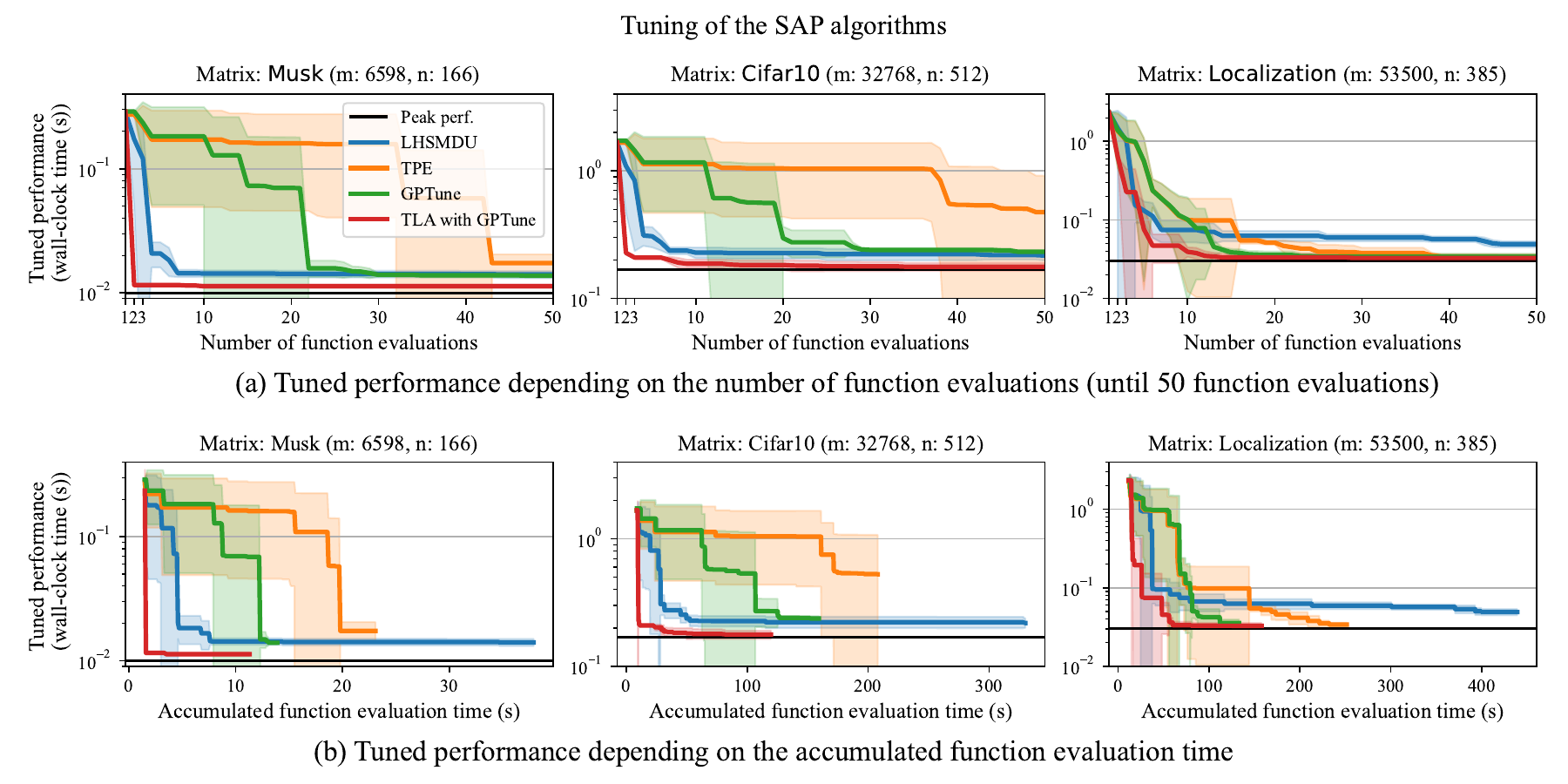}
    \centerline{\small (a) Hardly constrained $\textit{ARFE}$: $\textit{penalty\_factor}=5.0$, $\textit{allowance\_factor}=2.0$, $\textit{safety\_factor}=4$.}
    \includegraphics[width=\linewidth]{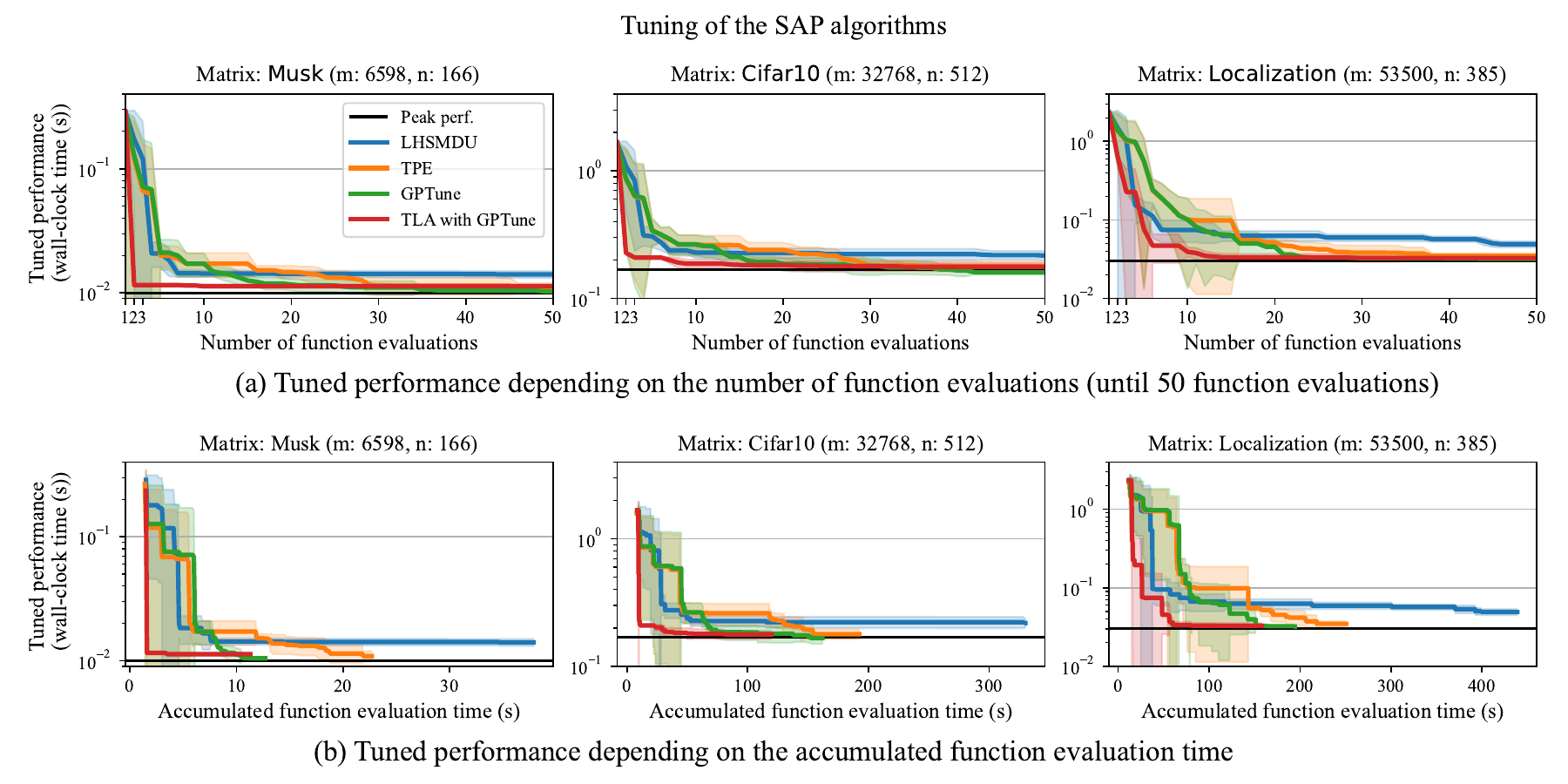}
    \centerline{\small (b) Softly constrained $\textit{ARFE}$: $\textit{penalty\_factor}=2.0$, $\textit{allowance\_factor}=100.0$, and $\textit{safety\_factor}=0$.}
    \caption{Tuning results for the three real-world input matrices for varying user-given constant parameters for $\textit{penalty\_factor}$, $\textit{allowance\_factor}$ and $\textit{safety\_factor}$.}
    \label{fig:tuning_results_real_world_soft_strong}
\end{figure}
}

\newpage

\end{document}